\documentclass[10pt,twocolumn,letterpaper]{article}

\usepackage[pagenumbers]{cvpr} 

\newcommand{\model}{DocSLM}

\usepackage{makecell}

\usepackage{booktabs}       
\usepackage{multirow}       
\usepackage{colortbl}       

\usepackage{arydshln}       
\usepackage{graphicx}       
\usepackage{caption}        
\usepackage{graphicx} 
\usepackage{colortbl} 
\usepackage{multirow} 
\usepackage{graphicx}       
\usepackage{array}          

\definecolor{cvprblue}{rgb}{0.21,0.49,0.74}
\usepackage[pagebackref,breaklinks,colorlinks,allcolors=cvprblue]{hyperref}
\usepackage{multirow}
\usepackage[ruled,vlined]{algorithm2e}
\SetKwComment{Comment}{/* }{ */}
\SetKwInput{KwIn}{Input}
\SetKwInput{KwOut}{Output}
\def\paperID{17410} 
\def\confName{CVPR}
\def\confYear{2026}

\title{DocSLM: A Small Vision-Language Model \\ for Long Multimodal Document Understanding}

\author{
Tanveer Hannan$^{1,2,3}$\thanks{Work done during an internship at Microsoft.\\ Correspondence to: hannan@dbs.ifi.lmu.de} \quad
Dimitrios Mallios$^{1}$ \quad
Parth Pathak$^{4}$ \quad
Faegheh Sardari$^{1}$ \\
Thomas Seidl$^{2,3}$ \quad
Gedas Bertasius$^{5}$ \quad
Mohsen Fayyaz$^{1}$ \quad
Sunando Sengupta$^{1}$ \\
$^1$ Microsoft \quad
$^2$ LMU Munich \quad
$^3$ MCML \quad
$^4$ FAIR Meta \quad
$^5$ UNC Chapel Hill \quad
}
\begin{document}

\maketitle
\begin{abstract}
Large Vision–Language Models (LVLMs) have demonstrated strong multimodal reasoning capabilities on long and complex documents. However, their high memory footprint makes them impractical for deployment on resource-constrained edge devices. We present \model, an efficient Small Vision–Language Model designed for long document understanding under constrained memory resources. \model~ incorporates a Hierarchical Multimodal Compressor that jointly encodes visual, textual, and layout information from each page into a fixed-length sequence, greatly reducing memory consumption while preserving both local and global semantics. To enable scalable processing over arbitrarily long inputs, we further introduce a Streaming Abstention mechanism that operates on document segments sequentially and filters low-confidence responses through an entropy-based uncertainty calibrator. Across multiple long multimodal document benchmarks, \model~matches or surpasses state-of-the-art methods while using 82\% fewer visual tokens, 75\% fewer parameters, and 71\% lower latency—delivering reliable multimodal document understanding on lightweight edge devices. Code and Model are available in \href{https://github.com/Tanveer81/DocSLM.git}{https://github.com/Tanveer81/DocSLM.git}.
\end{abstract}
    
\section{Introduction}
\label{sec:intro}
\begin{figure}[!t]
    \centering
    \includegraphics[width=\linewidth]{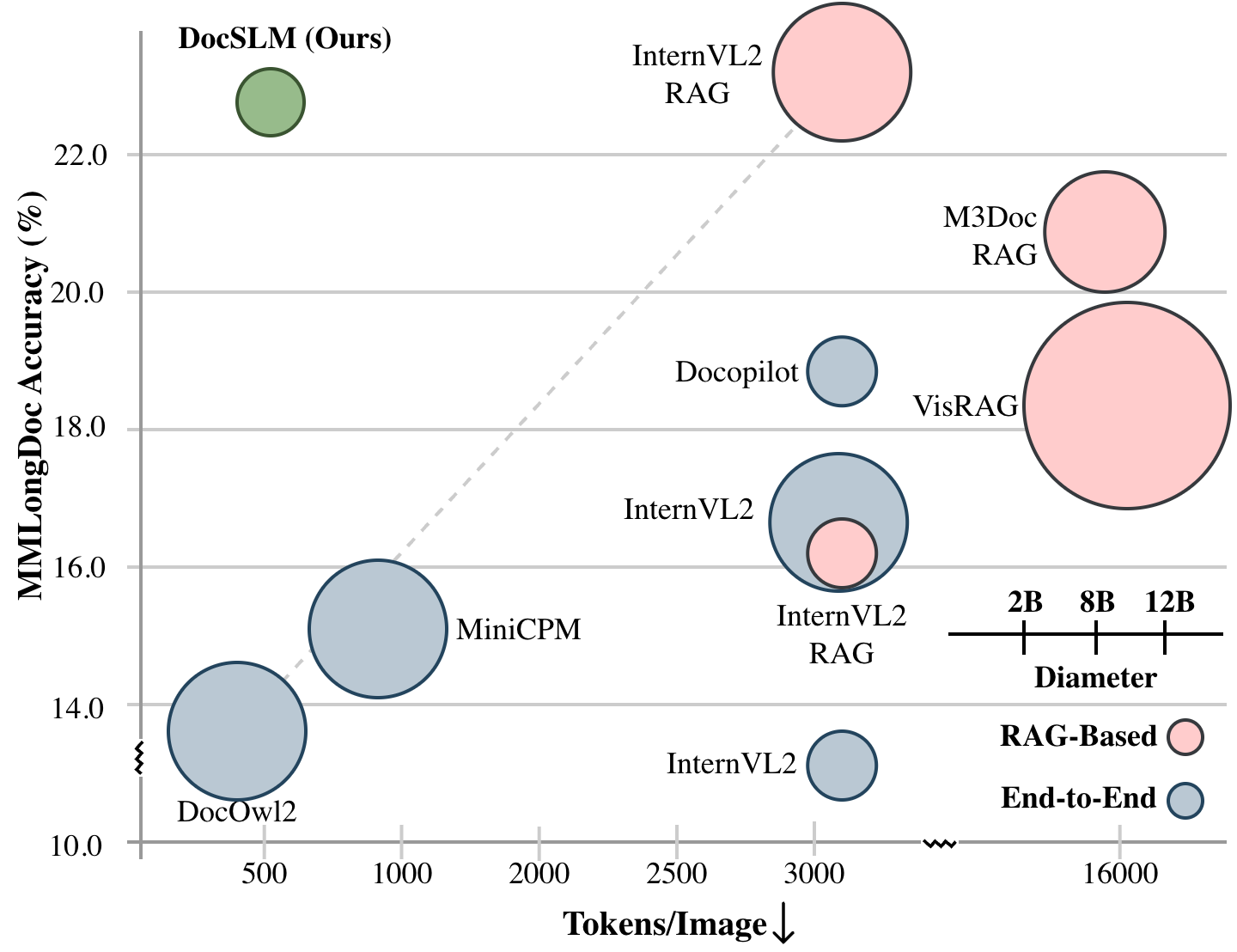}
    \vspace{-5mm}
    \caption{\textbf{Model accuracy versus token efficiency} on MMLongDoc~\cite{ma2024mmlong}. 
    \model~achieves a +9.3\% gain over DocOwl2~\cite{hu2024mplugdocowl2} with a comparable Tokens/Image budget, 
    while using 75\% fewer parameters than large RAG-based models such as InternVL2-RAG~\cite{wang2024needle}, 
    and outperforming the similarly sized Docopilot-2B~\cite{duan2025docopilot} by +0.9\% despite its significantly larger token budget.
    }
    \label{fig:latency_acc}
    \vspace{-7mm}
\end{figure}

Large Vision-Language Models (LVLMs) have made remarkable progress in understanding multimodal documents that integrate text, figures, and visual elements~\cite{duan2025docopilot, nacson2024docvlmmakevlmefficient, hu2024mplugdocowl2}. These capabilities are particularly important for understanding financial and technical reports, industrial documents, presentation slides, and scientific papers.
Despite recent progress, scaling vision–language models to longer context lengths remains a fundamental challenge, especially under constrained memory resources. LVLMs frequently exceed 8B parameters and can reach the total memory capacity of typical edge GPUs~\cite{tang2025scaling} during inference (Fig.~\ref{fig:memory}), which makes deployment on mobile or embedded devices particularly difficult~\cite{zhu2025simpleeffectivelayouttoken,chen2024far,duan2025docopilot,idefics,hu2024mplugdocowl2,longva,llava-next-interleave,nacson2024docvlmmakevlmefficient}. However, model size is only part of the cost: the number of input tokens greatly increases computational and memory demands. Document-focused LVLMs must process tens of pages rather than a single image, and each page can contain dense text, tables, figures, and complex layouts that inflate token counts. The burden rises further when systems incorporate OCR signals to read textual content, since many recent approaches feed OCR cues alongside visual tokens into the model~\cite{guan2025token,xiao2025adaptivemarkuplanguagegeneration,yu2025docthinkerexplainablemultimodallarge,zhu2025simpleeffectivelayouttoken}. OCR-enhanced methods improve accuracy in reading documents but also amplify memory and compute requirements, compounding the challenge of efficient scaling and practical deployment on resource-constrained hardware.

To minimize input context, recent Retrieval-Augmented Generation (RAG) frameworks, such as ~\cite{yu2024visrag,cho2024m3docrag,wang2024needle,chen2024sv,tanaka2025vdocragretrievalaugmentedgenerationvisuallyrich}, shorten document length by retrieving only the most relevant pages.
However, RAG methods often rely on segmented document retrieval and multi-stage query pipelines, which fragment contextual information and introduce additional retrieval latency~\cite{duan2025docopilot}. InternVL2-RAG~\cite{wang2024needle} exhibits a token generation latency of 113.4,ms, which is 3.5$\times$ slower than compact non-RAG models~\cite{duan2025docopilot,InternVL2}, and also incurs significantly higher memory usage (Fig.~\ref{fig:memory}), thereby limiting its practicality for interactive or edge-device scenarios.
 
To reduce memory consumption, recent methods use Smaller Vision–Language Models (SVLMs)~\cite{yu2025docthinkerexplainablemultimodallarge,huang2024minimonkey,chen2024far,duan2025docopilot}. Although they save memory by reducing parameter counts, they still depend on dense visual encodings (3K–9K tokens per document page) to compensate for limited modeling capacity.
Such high token counts quickly exceed the context length and memory capacity of mobile or edge devices, hindering scalability to multi-page documents. For instance, the state-of-the-art small model Docopilot-2B~\cite{wang2024needle} requires about 3,133 tokens per image, which nearly saturates the input capacity of edge GPUs (1,440–4,320 tokens) and thus accommodates only a single image per inference, making multi-page document processing infeasible. On the other hand, DocOwl2~\cite{hu2024mplugdocowl2} offers the lowest token count per image but degrades performance due to over-aggressive compression (Fig. \ref{fig:latency_acc}).

To this end, we introduce \textbf{\model~}, a lightweight Vision–Language Model designed for reliable long-document understanding under strict memory and input token-length constraints.
Documents naturally consist of sequences of pages, each containing multimodal information. Consequently, addressing the memory challenges of document-understanding VLMs requires solutions at both the page level and the document level.
At the page level, we propose a \textbf{Hierarchical Multimodal Compression} module that jointly encodes visual, textual, and layout features from each document page into a \textbf{fixed 576 tokens}—independent of the number of OCR tokens—while preserving fine-grained semantic and spatial information. As shown in Fig. \ref{fig:memory}, this compression method substantially reduces memory overhead compared to similar sized models Docopilot-2B~\cite{duan2025docopilot} and InternVL2-2B~\cite{InternVL2}. 

However, reducing per-page memory footprint alone is not sufficient. Even with efficient page-level encoding, memory usage still grows rapidly as the number of pages increases (Fig.~\ref{fig:memory}). To address this document-level challenge, we propose a \textbf{Streaming Abstention} mechanism that processes long documents sequentially in a segment-wise manner. 
The input document is divided into segments, each independently encoded to produce an intermediate prediction along with an uncertainty score. This ensures a constant memory footprint regardless of document length.
While each segment is encoded independently to maintain a constant memory footprint, \model~preserves contextual continuity through a streaming mechanism that implicitly carries information across segments via textual cues and the model’s calibrated uncertainty.
This allows the model to achieve full-document understanding without storing cross-segment activations.

With this design, \model~can process up to 120-page documents from MMLongDocBench~\cite{ma2024mmlong} with a \textbf{$\sim$14 GB peak GPU memory}. Finally, an uncertainty calibrator aggregates all valid segment-level predictions and selects the most reliable document-level answer based on uncertainty. Together, these components enable \model~to handle arbitrarily long documents efficiently under limited GPU or edge-device memory, while providing stable and accurate document-level understanding across segments. Our contributions can be summarized as follows:

\begin{figure}[t]
\centering
\includegraphics[width=.9\linewidth]{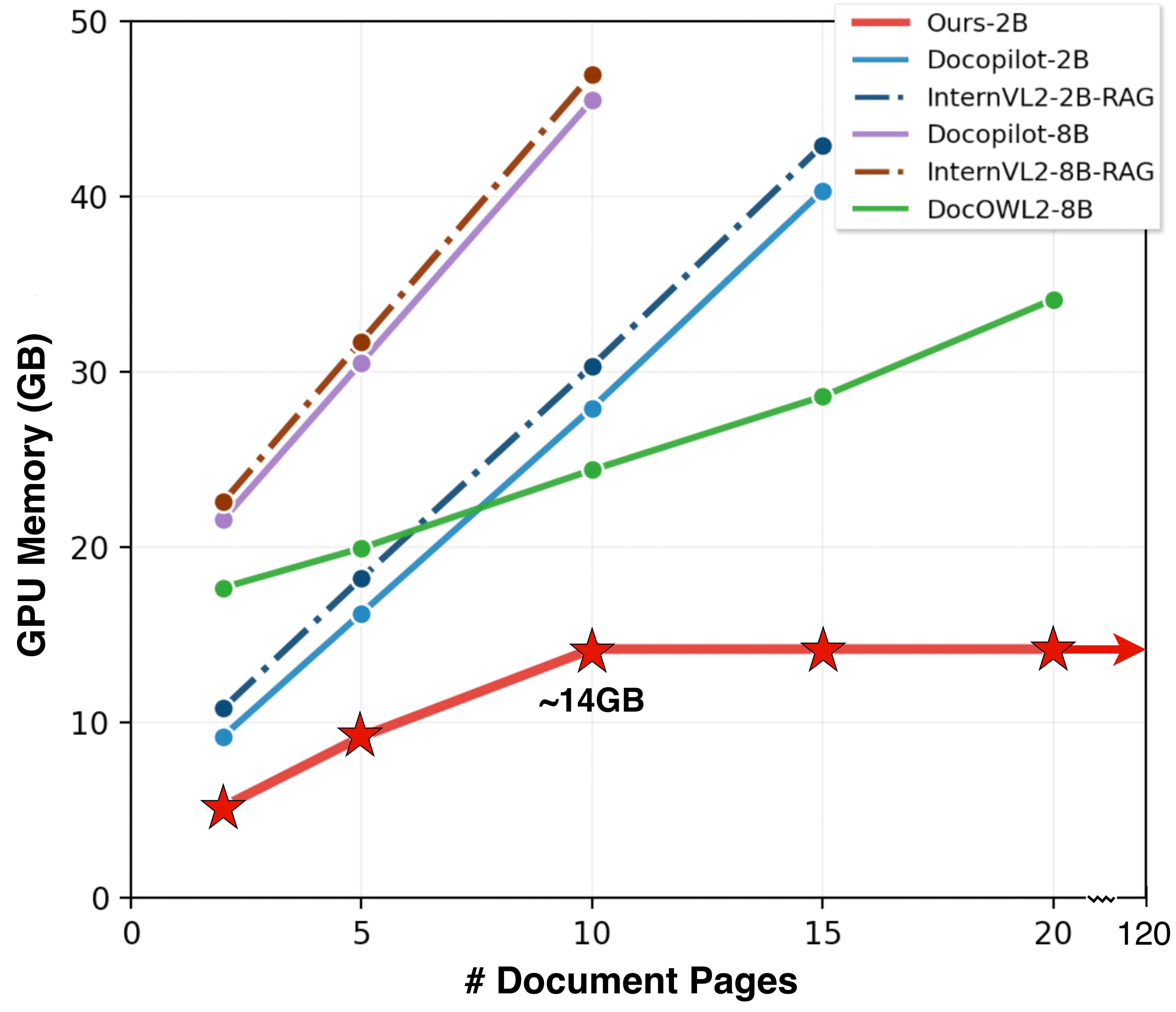}
\vspace{-5mm}
\caption{\textbf{Peak GPU memory usage vs.\ number of document pages.}
We evaluate all models using their official implementations under identical inference settings, progressively increasing the number of document pages. Peak GPU memory is measured using PyTorch profiling tools.
\model~achieves the lowest memory footprint among all methods, maintaining a constant plateau of \textbf{$\sim$14 GB} beyond 10 pages due to its streaming mechanism, enabling scalable inference on resource-constrained devices. 
}
\label{fig:memory}
\vspace{-7mm}
\end{figure}
\begin{itemize}
    \item We introduce \model~, a compact 2B-parameter Vision–Language Model that uses \textbf{82\%} fewer visual tokens and \textbf{75\%} fewer parameters than existing large LVLMs and retrieval-augmented models.
    \item We propose a Hierarchical Multimodal Compression module that achieves a \textbf{5.6$\times$} reduction in input tokens by jointly encoding visual, textual, and layout features, while a Streaming Abstention mechanism maintains a constant memory footprint—enabling efficient inference over arbitrarily long documents across edge devices. 
    \item Despite its compact size, \model~achieves state-of-the-art performance, surpassing DocOwl2-8B~\cite{hu2024mplugdocowl2} by \textbf{+9.3\%} under a comparable token budget and outperforming the similarly sized Docopilot-2B~\cite{duan2025docopilot} by \textbf{+0.9\%}, while reducing latency by \textbf{3.5$\times$} (32.1\,ms vs.\ 113.4\,ms) compared to InternVL2-RAG~\cite{wang2024needle}. 
\end{itemize}
\section{Related Works}
\label{sec:related}
\noindent\textbf{Document Understanding.}   
OCR-free models~\cite{li2023blip,alayrac2022flamingo,liu2024llavanext,InternVL2} typically process high-resolution document images or tiled patches to capture global visual context but often struggle with densely packed text regions. On the other hand, OCR-enhanced approaches~\cite{wang2023docllm,blau2024gram,biten2022latr,ganz2023towards,ye2023deepsolo} extract textual content and layout information, resulting in long input sequences, particularly for a multi-page document. 
Recent hybrid methods~\cite{guan2025token,yu2025docthinkerexplainablemultimodallarge} embed OCR text as language tokens for structured multimodal alignment, while layout-aware models~\cite{xiao2025adaptivemarkuplanguagegeneration,liao2025doclayllmefficientmultimodalextension,wang2025martenvisualquestionanswering} explicitly encode markup or spatial structures to enhance localized reasoning.  
Our main goal is to build an efficient document undesrstanding model that can run on resource-constrained edge devices.

\noindent\textbf{Document Compression.}  
OCR-free models~\cite{hu2024mplugdocowl2,alayrac2022flamingo,hu2024mplug,li2024tokenpacker} focus on specialized visual token compression but often fail to preserve fine-grained textual and layout cues essential for understanding text-heavy documents.
In OCR-based paradigms, GRAM~\cite{blau2024gram} introduces a Compression Transformer to aggregate OCR tokens across pages, improving long-document understanding at the cost of substantial model complexity.
DocVLM~\cite{nacson2024docvlmmakevlmefficient} encodes textual semantics into fixed-size OCR embeddings but still exhibits {linear token growth} as the number of pages increases.
In contrast, our hierarchical compression maintains a {constant token count per page} and does not incur any additional tokens from OCR inclusion, enabling scalable multimodal encoding for long documents.

\noindent\textbf{Long Multimodal Document Understanding.} 
In addition to the increasing of input tokens size, long document understanding poses additional challenges due to complex inter-page dependencies.  
One line of existing approaches tackles this problem through long-context vision–language models~\cite{zhu2025simpleeffectivelayouttoken,chen2024far,duan2025docopilot,idefics,hu2024mplugdocowl2,longva,llava-next-interleave,nacson2024docvlmmakevlmefficient}.  
For example, DocVLM~\cite{nacson2024docvlmmakevlmefficient} mitigates input redundancy using fixed-size OCR embeddings, LayTokenLLM~\cite{zhu2025simpleeffectivelayouttoken} encodes layout-aware OCR tokens without positional extrapolation, and Docopilot~\cite{duan2025docopilotimprovingmultimodalmodels} fine-tunes off-the-shelf LVLMs on large-scale instruction datasets.  
Meanwhile, RAG-based methods~\cite{yu2024visrag,cho2024m3docrag,wang2024needle,chen2024sv,tanaka2025vdocragretrievalaugmentedgenerationvisuallyrich} retrieve relevant document pages or visual embeddings before generation, but introduce additional retrieval latency and still require thousands of input tokens per page—restricting scalability to long documents.  
We propose a streaming model to handle long documents with a constant input token and memory footprint.
\section{Method}
\label{sec:method}
\begin{figure*}[!t]
    \centering
    \includegraphics[width=1\linewidth]{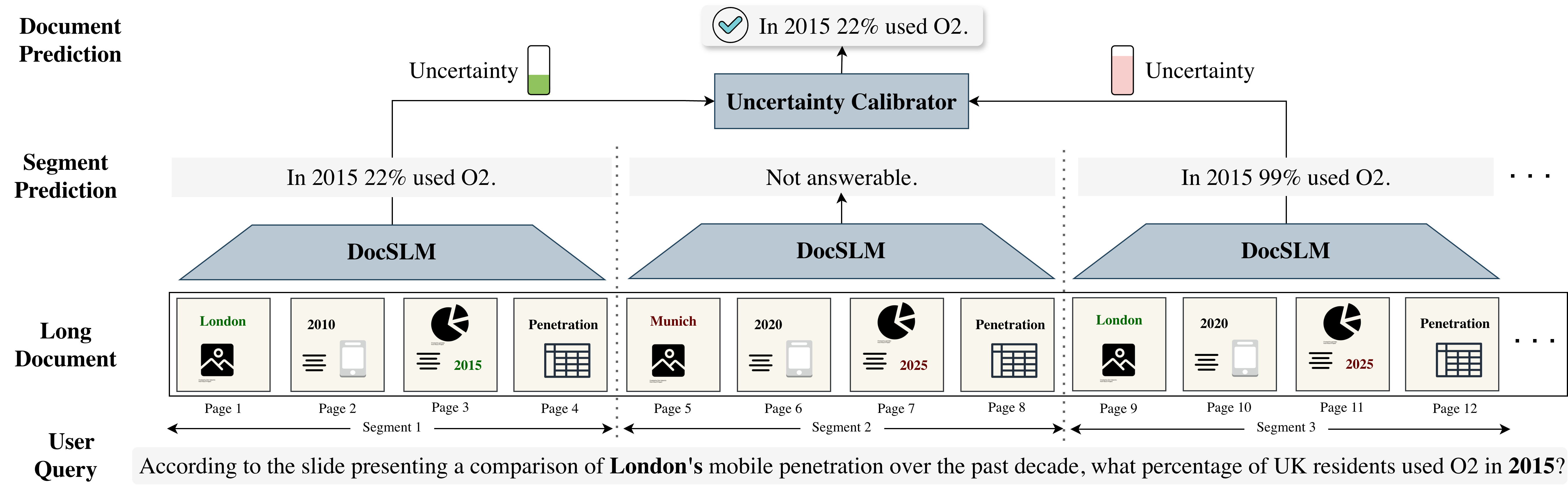}
    \vspace{-5mm}
   \caption{\textbf{Streaming Abstention.} 
    To process long documents efficiently, \model~divides the input into shorter segments that can be handled sequentially or in parallel. 
    Each segment produces an intermediate prediction that can either (\textbf{Left}) correctly answer the query with low uncertainty, (\textbf{Middle}) abstain when the answer is not present, or (\textbf{Right}) produce an incorrect answer with high uncertainty. 
    A \textit{Uncertainty Calibrator} aggregates all valid segment predictions and selects the final document-level answer corresponding to the lowest uncertainty. 
    In memory-limited settings, segments are processed sequentially, with memory from previous segments released before processing the next one. 
    }\vspace{-5mm}
    \label{fig:streaming}
\end{figure*}

Given a long multimodal document 
$\mathcal{D} = \{d^{1}, d^{2}, \dots, d^{N}\}$ with $N$ pages where $d^{n}$ is the $n^{th}$ page and a natural language query $S$, our goal is to predict a response $y$ that is consistent with the entire input. 
To achieve this under strict memory and context-length constraints, 
\model~introduces two key components:
(1) a {Hierarchical Multimodal Compression} module that condenses visual, textual, and layout features into a compact token representation per page, and 
(2) a {Streaming Abstention} mechanism that enables reliable reasoning over arbitrarily long inputs. 
Overview of the method is in Fig.~\ref{fig:streaming}.

\subsection{Hierarchical Multimodal Compression}
\begin{figure}[!t]
    \centering
    \includegraphics[width=.65\linewidth]{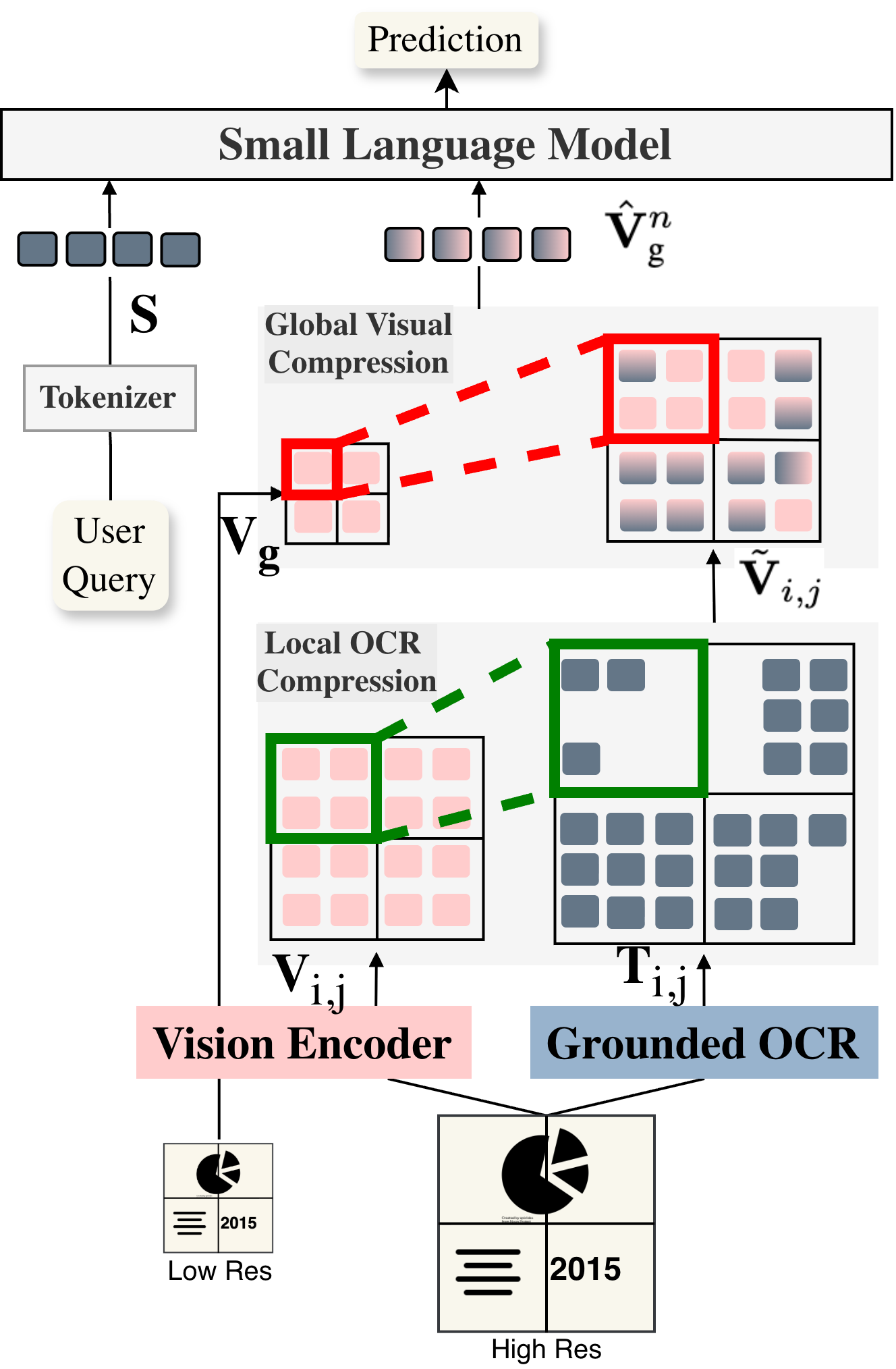}
    \vspace{-2mm}
    \caption{\textbf{Hierarchical Multimodal Compressor.}
    The Vision Encoder produces global ($\mathbf{V}_{g}$) and local ($\mathbf{V}_{i,j}$) visual features, while the Grounded OCR module provides region-aligned text embeddings ($\mathbf{T}_{i,j}$). 
    (\textbf{Bottom}) As indicated by \textcolor[HTML]{2CA02C}{green} boxes and dotted links, \textit{Local OCR Compression} performs spatially localized cross-attention—each visual patch $\mathbf{V}_{i,j}$ attends only to its paired OCR tokens $\mathbf{T}_{i,j}$—yielding compressed local features $\hat{\mathbf{V}}_{i,j}$. 
    (\textbf{Top}) The \textit{Global Visual Compression}, shown with dotted \textcolor[HTML]{D62728}{red} connections, aggregates these local representations by allowing the global visual feature $\mathbf{V}_{g}$ to attend selectively to compressed local regions $\hat{\mathbf{V}}_{i,j}$, producing the final global representation $\hat{\mathbf{V}}^{n}_{g}$.
    }
    \label{fig:compressor}
    \vspace{-5mm}
\end{figure}

Our compression module performs structured token reduction through a two-stage fusion process (Fig.~\ref{fig:compressor}).
In the first stage, local OCR compression aligns each visual region with its corresponding OCR text and layout using localized attention, merging them into compact region-level features.
In the second stage, global visual compression aggregates these regional features into a fixed-length page representation that preserves both spatial alignment and overall document semantics.

\noindent\textbf{Multimodal Feature Extraction.} The process starts with multimodal feature extraction. Specifically, for each document page $d^{n}$, we divide the image into a grid of $R \times C$ spatial crops
\begin{equation}
\mathcal{C}_{\text{loc}}^{n} = \{d_{i,j}^{n}\}_{i=1..R,\, j=1..C}
\end{equation}
and obtain a downsampled global crop $d_{\text{g}}^{n}$. Then, a shared vision encoder $E_v(\cdot)$ extracts patch-level visual features:
\begin{equation}
\mathbf{V}_{i,j}^{n} = E_v(d_{i,j}^{n}), \qquad 
\mathbf{V}_{\text{g}}^{n} = E_v(d_{\text{g}}^{n}).
\end{equation}

To incorporate textual cues, a lightweight OCR module produces $K^{n}$ word--bounding-box pairs 
$\{(s_k^{n}, b_k^{n})\}_{k=1}^{K^{n}}$. 
Each word token is embedded as 
$\mathbf{t}_k^{n} = E_t(s_k^{n})$ 
using the tokenizer of the Small Language Model (SLM). 
We then spatially associate OCR tokens with their corresponding image crops by bounding-box overlap:
\begin{equation}
\mathbf{T}_{i,j}^{n} = \{\mathbf{t}_k^{n} \mid \mathrm{IoU}(b_k^{n}, \mathrm{bbox}(I_{i,j}^{n})) > \tau \}
\end{equation}
where $\tau$ is a fixed overlap threshold. 
This mapping yields region-aligned OCR sets $\mathbf{T}_{i,j}^{n}$ for each visual crop $d_{i,j}^{n}$, ensuring that subsequent local compression attends only to semantically and spatially relevant text regions.

\noindent\textbf{Local OCR Compression.}
At the local level, visual and text features within each region $(i,j)$ are fused using cross-attention (CA):

\begin{equation}
\tilde{\mathbf{V}}_{i,j} = \mathbf{V}_{i,j} + \text{CA}\big(\mathbf{V}_{i,j}, \mathbf{T}_{i,j}\big)
\end{equation}
where $\mathbf{V}_{i,j}$ serves as queries and $\mathbf{T}_{i,j}$ as keys and values. 
This enriches local visual tokens with corresponding OCR semantics without increasing the overall sequence length, achieving spatially aligned multimodal fusion.

\noindent\textbf{Global Visual Compression.}
The local features $\tilde{\mathbf{V}}{i,j}$ are spatially aligned and preserve high-resolution details, but they result in long token sequences.
To reduce the total number of tokens, these local representations $\tilde{\mathbf{V}}_{i,j}$ are summarized into compact global features $\mathbf{V}{\text{g}}$ through an additional cross-attention layer:
\begin{equation}
\hat{\mathbf{V}}_{g,i,j} = \mathbf{V}_{\text{g},i,j} + \text{CA}(\mathbf{V}_{\text{g},i,j}, \tilde{\mathbf{V}}_{i,j})
\end{equation}
producing compact, spatially consistent representations $\hat{\mathbf{V}}_{i,j}$. {Subsequently,} all regional features are concatenated to form the final page-level representation:
\begin{equation}
\hat{\mathbf{V}}_{\text{g}}^{n} = \text{Concat}(\{\hat{\mathbf{V}}_{g,i,j}\}_{i=1..R,\,j=1..C})
\end{equation}
This hierarchical compression ensures that each page—regardless of OCR token count—is represented by a fixed number of tokens while preserving both local fine-grained and global structural details.

\begin{table*}[!ht]
\centering
\scriptsize
\setlength{\tabcolsep}{6pt}
\renewcommand{\arraystretch}{1.12}
\vspace{1mm}
\resizebox{\textwidth}{!}{
\begin{tabular}{ll cc}
\toprule
\textbf{Stage} & \textbf{Curriculum Goal} & \textbf{Trainable Modules} & \textbf{Dataset Source}~\cite{hu2024mplug_docowl_1_5,hu2024mplugdocowl2} \\
\midrule
Pretrain~1 & Image–OCR Alignment & OCR Compressor & DocStruct4M (25\%) \\
Pretrain~2 & Image Compression & Vision Encoder, OCR \& Vision Compressor & DocStruct4M (Remaining 75\%) \\
Pretrain~3 & Document Compression & OCR \& Vision Compressor & DocStruct4M (Random 25\%), MP-DocStruct1M \\
\midrule
Finetune~1 & Instruction Following & OCR \& Vision Compressor, SLM & DocDownstream~1.0 \\
\rowcolor{gray!8}Finetune~2 & Streaming Abstention & OCR \& Vision Compressor, SLM & DocDownstream~2.0, DocGenome12K, MP-DocReason51K \\
\bottomrule
\end{tabular}
}
\vspace{-3mm}
\caption{\textbf{Curriculum Training Stages.}
The model is progressively trained from single-page pretraining to multi-document finetuning, with increasingly complex objectives and negative-pair supervision. An MLP adapter is trained in all stages.}
\label{tab:train_config_modules}
\vspace{-5mm}
\end{table*}

\subsection{Streaming Abstention}
\label{subsec:streaming}

Even after hierarchical multimodal compression, extremely long documents can still exceed device memory limits during inference. 
To address this, we introduce the {Streaming Abstention} mechanism (Fig.~\ref{fig:streaming}), which enables us to process {arbitrarily long inputs under constant GPU memory usage}. 

\noindent\textbf{Document Segmentation.}
For a sequence of length $N$, attention memory scales linearly with sequence length, i.e., $\mathcal{O}(N)$ per layer when using optimized kernels such as FlashAttention~\cite{dao2022flashattention}.
However, even linear growth becomes impractical on memory-constrained edge devices. To reduce the peak memory usage at any given time, we divide the document into $T$ smaller segments of equal length $N/T$ denoted as $\{s_t\}_{t=1}^{T}$.

\noindent Each segment can be processed with $\mathcal{O}(N/T)$ memory, reducing the peak GPU usage by roughly $T\times$ when processed sequentially. 
This segmentation strategy enables inference over extremely long sequences while keeping memory within device constraints (refer Tab.~\ref{tab:module}). 
While segment-wise processing guarantees constant-memory inference, it still requires an aggregation mechanism to coherently integrate information across segments.
We therefore propose an {uncertainty-guided aggregation} approach that fuses the most confident segment outputs into a coherent document-level prediction—achieved in a single forward pass without any additional model calls.

\noindent\textbf{Segment-Wise Processing.}
Unlike traditional streaming models that retain key–value (KV) caches across segments~\cite{di2025streaming,xiao2023efficient}, \model~stores only the textual prediction and the SLM’s intrinsic uncertainty for each segment, avoiding large memory accumulation. 
For each segment $\mathcal{S}_t$, the Hierarchical Multimodal Compressor produces compressed embeddings:
\begin{equation}
\hat{\mathbf{V}}_{s_t} = \{\hat{\mathbf{V}}_{\text{g}}^{n} \mid d^{n} \in s_t\}
\end{equation}
which are fed into the SLM together with the query $S$ to generate a segment-level prediction:
\begin{equation}
p_t = \text{SLM}(\hat{\mathbf{V}}_{s_t}, S)
\end{equation}
Before releasing activation memory, \model~estimates the predictive uncertainty for each segment using token-level entropy, where $u_t$ denotes the average uncertainty of the generated text distribution for segment $s_t$.
\begin{equation}
u_t = -\frac{1}{T}\sum_{k=1}^{T}\sum_{w} p(w \mid T_{<k}, s_t)
\log p(w \mid T_{<k}, s_t)
\end{equation}
After storing the prediction text and its corresponding uncertainty, all intermediate activations and KV caches from $s_t$ are released, maintaining a constant GPU memory.

\noindent\textbf{Uncertainty-Based Aggregation.}
Among the valid segment predictions $\mathcal{P}_{\text{valid}}$, the final document-level answer is obtained by selecting the most confident one:
\begin{equation}
\hat{y} = \arg\min_{p_t \in \mathcal{P}_{\text{valid}}} u_t
\end{equation}
\model~ produces calibrated uncertainty estimates through its learned abstention mechanism during Finetuning Stage~2 (Sec.~\ref{subsec:training}). It enables robust evidence aggregation across arbitrarily long documents.
Notably, sequential processing does not compromise accuracy; in fact, the uncertainty-guided aggregation enhances performance by emphasizing confident evidence (Tab.~\ref{tab:module}). We also explored hierarchical aggregation, where high-confidence predictions are reused as input to the SLM. 
While yielding slight accuracy gains, these methods incurred extra computation and latency, making them less suitable for edge deployment. 
Hence, we adopt the single-pass uncertainty-guided selection for its balance of reliability and efficiency.

\subsection{Training}
\label{subsec:training}
Tab.~\ref{tab:train_config_modules} summarizes the multi-stage curriculum used to train \model.
The training progresses from low-level single-page pretraining to high-level multi-document finetuning with gradually increasing task complexity.

\noindent\textbf{Pretraining Stages} focus on multimodal representation learning. In {Pretraining~1}, only the OCR Compressor is optimized to align textual embeddings with their corresponding visual regions. 
{Pretraining~2} extends training to the Vision Encoder and both OCR and Vision Compressors, using single-page documents to learn consistent visual–text fusion. Finally, {Pretraining~3} introduces multi-page documents, encouraging the model to encode coherent representations across multiple visual contexts.

\noindent\textbf{Finetune Stage~1} serves as the first step in training the Small Language Model (SLM). Through instruction tuning on single-image inputs, the model learns to interpret multimodal cues and follow natural language instructions, establishing a foundation for subsequent multi-image and long-document understanding.

\noindent\textbf{Finetune Stage 2.}  
This stage extends training to multi-image and multi-document settings with \textit{negative-pair supervision}, where question–answer pairs are randomly mismatched with unrelated document segments to simulate incomplete or irrelevant context. 
For each positive segment $s_t$ containing evidence for a query $q$, a negative counterpart $s_t^{-}$ is constructed from unrelated content where $q$ cannot be answered. 
\model~is trained to detect such mismatches and \textit{abstain} from unsupported predictions by generating an explicit \textit{``Not Answerable''} token sequence. 
This dual supervision over $\{s_t, s_t^{-}\}$ calibrates model confidence between valid and invalid contexts, ensuring reliable streaming inference under uncertain or partial evidence.

\noindent Across all stages, the model is trained using a standard next-token prediction loss on instruction-tuned triplets $\{(s_t, q, y)\}$, where $s_t$ denotes the input segment, $q$ the instruction query, and $y$ the target response:
\begin{equation}
\mathcal{L}_{\text{NTP}} = - \sum_{w} \log p_\theta(w \mid T_{<k},\, \mathbf{X}(s_t), q)
\end{equation}
This curriculum progressively transitions \model~from learning localized visual–text compression to performing robust, long-context multimodal understanding.

\begin{table*}[!ht]
\centering
\scriptsize
\setlength{\tabcolsep}{3.8pt}
\renewcommand{\arraystretch}{1.05}
\begin{tabular}{c lcccccccc}
\toprule
& \textbf{Model} &
\textbf{Tok/Image}$\downarrow$ &
\textbf{Param}$\downarrow$ &
\textbf{Latency (ms)$\downarrow$} &
\textbf{MMLDoc (Acc)$\uparrow$} &
\textbf{MP-DocVQA (ANLS)$\uparrow$} &
\textbf{DUDE (ANLS)$\uparrow$} &
\textbf{NewsVQA (ANLS)$\uparrow$} \\
\midrule
\multirow[c]{8}{*}{\rotatebox[origin=c]{90}{\parbox{1.8cm}{\centering\textbf{Large}\\\textbf{Models}}}} 
 & LayTokenLLM~\cite{zhu2025simpleeffectivelayouttoken}              & Var.  & 8B   & --    & --          & 74.3 & \textbf{52.0} & -- \\
& InternVL2~\cite{chen2024far}                & $\sim$3,133        & 8B   & 81.0  & \underline{17.4} & 79.3 & 37.0 & 53.0 \\
& Docopilot~\cite{duan2025docopilot}                & $\sim$3,133        & 8B   & 81.0  & \textbf{28.8} & \underline{81.3} & -- & -- \\
& Idefics3~\cite{idefics}                 & 838       & 8B   & --    & --          & 67.2 & 38.7 & \textbf{60.2} \\
& DocOwl2~\cite{hu2024mplugdocowl2}                   & \textbf{324} & 8B & --    & 13.4        & 69.4 & 46.8 & -- \\
& LongVA~\cite{longva}                   & $\sim$2,029        & 7B   & --    & --          & 60.8 & 38.4 & 50.6 \\
& LLaVA-Next~\cite{llava-next-interleave}  & \underline{729}       & 7B   & --    & --          & 44.9 & 28.0 & \underline{56.7} \\
& DocVLM~\cite{nacson2024docvlmmakevlmefficient}                   & 1088      & 7B   & --    & --          & \textbf{84.5} & \underline{47.4} & -- \\
\midrule

\multirow[c]{5}{*}{\rotatebox[origin=c]{90}{\parbox{1.6cm}{\centering\textbf{RAG}\\\textbf{Models}}}} 
& VisRAG~\cite{yu2024visrag}                   & Var.  & 12B  & 288.3 & 18.8        & --   & --   & 36.3 \\
& InternVL2+RAG~\cite{wang2024needle}             & $\sim$3,133        & {8B}   & \underline{113.4}  & 24.2 & \underline{78.7} & -- & -- \\
& M3DocRAG~~\cite{cho2024m3docrag}                 & 16,384       & 7B   & --    & \underline{21.0} & \textbf{84.4} & -- & -- \\
& SV-RAG~\cite{chen2024sv}                   & \underline{3,072}        & 4B   & --    & \textbf{23.0} & 71.0 & \underline{45.0} & \textbf{61.0} \\
& VDocRAG~\cite{tanaka2025vdocragretrievalaugmentedgenerationvisuallyrich}                  & \textbf{768} & {4B} & --    & --          & --   & \textbf{48.5} & \underline{44.2} \\
& InternVL2+RAG~\cite{wang2024needle}             & $\sim$3,133        & {2B}   & \textbf{82.9}  & 17.2 & {72.6} & -- & -- \\
\midrule

\multirow[c]{5}{*}{\rotatebox[origin=c]{90}{\parbox{1cm}{\centering\textbf{Small}\\\textbf{Models}}}} 
& DocThinker~\cite{yu2025docthinkerexplainablemultimodallarge}               & 9,216        & 3B   & --    & --          & --   & \underline{21.3} & -- \\
& MiniMonkey~\cite{huang2024minimonkey}               & \underline{3,072}        & 2B   & --    & 10.3        & 70.3 & -- & -- \\
& InternVL2~\cite{chen2024far}                 & $\sim$3,133        & 2B   & \underline{35.9}   & 10.5 & \underline{71.8} & -- & -- \\
& Docopilot~\cite{duan2025docopilot}$^{\ast}$      & $\sim$3,133        & 2B   & \underline{35.9}  & \underline{21.8} & \textbf{76.2} & -- & -- \\
\rowcolor{gray!10}& \textbf{Ours} & \textbf{576} & 2B & \textbf{32.1} & \textbf{22.7} & 70.0 & \textbf{47.6} & \textbf{66.2} \\
\bottomrule
\end{tabular}
\caption{
\textbf{Main results on long-document benchmarks} across large-scale, retrieval-augmented, and compact vision-language models. 
$\text{“Var.”}$ denotes variable-length inputs without a fixed tokenization limit. 
For RAG-based models, token counts refer to the generator module only (excluding retriever overhead). 
Best and second-best results per column are shown in \textbf{bold} and \underline{underline}, respectively. Despite operating with only 576 tokens per image and 2B parameters, \textbf{\model~}matches or exceeds the performance of 7B–8B models and RAG-enhanced systems across most benchmarks, while achieving the lowest measured latency. 
}
\vspace{-5mm}
\label{tab:main}
\end{table*}

\section{Experimental Setup}
\subsection{\model\ Baselines}
Long-document understanding under memory constraints remains underexplored for Small Vision–Language Models (SVLMs). 
For fair comparison, we evaluate two backbone-consistent variants built on SigLIP2~\cite{tschannen2025siglip} and Qwen2.5~\cite{qwen2025qwen25technicalreport}, also used in our proposed DocSLM:

\noindent\textbf{OCR-Free Baseline.}  A LLaVA-Next–style~\cite{llava-next-interleave} model that interleaves visual and text tokens without explicit OCR fusion, relying solely on dense visual embeddings.

\noindent\textbf{OCR Baseline.}  Extends the above by incorporating OCR text from PaddleOCR~\cite{cui2025paddleocr}, serialized and appended to the visual tokens for joint visual–text attention. \model~builds upon the backbone of OCR-Free Baseline.

\subsection{Datasets and Metrics}

\noindent\textbf{MP-DocVQA Dataset}~\cite{mpdocvqa} comprises approximately 46K QA pairs derived from 60K scanned pages of around 6K industrial documents, encompassing tables, diagrams, figures, and both handwritten and printed text.  

\noindent\textbf{DUDE Dataset}~\cite{dude} expands the coverage to multiple real-world domains, including medical, legal, financial, and technical reports with 41K QA pairs across 5K documents.  

\noindent\textbf{MMLongBench-Doc Dataset}~\cite{ma2024mmlongbench} extends the evaluation scope by incorporating documents with considerably greater lengths, averaging 47.5 and a maximum of 120 pages per document, respectively. 

\noindent\textbf{NewsVideoQA Dataset}~\cite{newsvideoqa} focuses on text-rich broadcast videos collected from major global news outlets such as BBC and CNN, providing 8K QA pairs grounded in 3K news clips containing dynamic, text-heavy scenes.
Although primarily a video QA benchmark, its frames often contain overlaid text, captions, and layout-rich visual elements similar to real-world documents.
We include this dataset to evaluate the model’s ability to generalize its document understanding capabilities to temporally varying, text-centric visual content.

\noindent\textbf{Evaluation Metric.}  
We evaluate our model on the multimodal \textit{Document Question Answering} (DocQA) task. Following prior works, we use the \textit{Average Normalized Levenshtein Similarity} (ANLS)~\cite{biten2019scene} as the primary evaluation metric. ANLS computes the normalized edit similarity between predicted and ground-truth answers, averaged over all samples, with scores below a threshold of $\tau=0.5$ set to zero. For MMLongDocBench, which contains long-document question–answer pairs, the same thresholding rule is applied, but the metric reports binary \textit{Accuracy}.

\subsection{Implementation Details.}  
In our framework, we use SigLIP2~\cite{tschannen2025siglip} as the vision encoder, PaddleOCR~\cite{cui2025paddleocr} for OCR extraction, and Qwen2.5-1.5B~\cite{qwen2025qwen25technicalreport} as the Small Language Model (SLM). Training is performed using Fully Sharded Data Parallel (FSDP) across 8 nodes, each equipped with 4 NVIDIA A100 (80 GB) GPUs, resulting in a total of 32 GPUs. We use a total batch size of {1024} during pretraining and {256} during finetuning. Learning rate is set to $1e-4$ for pretraining and $2e-5$ for finetuning. Training steps count and batch size for each stage are listed in Tab. \ref{tab:train_stage}. We use the AdamW~\cite{loshchilov2019decoupledweightdecayregularization} optimizer with a cosine learning rate schedule and an initial warm-up phase. More details are in the Supplementary material \Cref{sec:sup implementation} and \Cref{sec:sup edge}.
\section{Experiments}
\label{sec:result}
\noindent\textbf{Comparison with Large and RAG Models.}  
Tab. ~\ref{tab:main} compares \model\ with large-scale and retrieval-augmented approaches.  
Large LVLMs, such as DocVLM-7B~\cite{nacson2024docvlmmakevlmefficient} (1{,}088 input tokens per image) and Docopilot-8B~\cite{duan2025docopilot} (3{,}133 input tokens per image) achieve strong accuracy but incur high memory requirements, with inference latencies of 81–113\,ms. 
RAG-based methods, including InternVL2+RAG~\cite{wang2024needle} and M3DocRAG~\cite{cho2024m3docrag} introduce additional retrieval overhead, often exceeding 110\,ms per sample.  
In contrast, \model\ operates with only \textbf{576} tokens per image—\textbf{5.4$\times$} fewer than large models—achieving 22.7\% on MMLDoc, 70.0 ANLS on MP-DocVQA, and 47.6 ANLS on DUDE at just \textbf{32.1\,ms latency}.  
This corresponds to a \textbf{+5.7 pp gain} over InternVL2-8B on DUDE and near-parity with the 8B Docopilot on MMLDoc despite using \textbf{75\% fewer parameters}.  
Even compared to 8B RAG models, \model\ retains over \textbf{95\%} of their accuracy while running \textbf{3.5$\times$ faster}, underscoring the efficiency of multimodal compression for resource-constrained reasoning.

\vspace{1mm}
\noindent\textbf{Comparison with Small Vision–Language Models.}  
Within the 2–3B parameter range, \model\ achieves the best trade-off between efficiency and accuracy (Tab. ~\ref{tab:main}).  
Compared to Docopilot-2B~\cite{duan2025docopilot} (3,133 tokens, 35.9\,ms), it runs faster (\textbf{32.1\,ms}) while improving by \textbf{+0.9 pp} on MMLDoc and \textbf{+26.3 pp} on DUDE.  
Relative to InternVL2-2B~\cite{chen2024far}, \model\ gains \textbf{+12.2 pp} on MMLDoc and \textbf{+47.6 pp} on DUDE, highlighting its superior multimodal reasoning and text–layout alignment under tight token budgets.

\subsection{Generalization to Video Question Answering.}  
Despite being trained mainly on documents, with only 8.6K video samples versus 6.75M document annotations, \model\ generalizes effectively, achieving state-of-the-art ANLS of \textbf{66.2} (Tab. ~\ref{tab:main}).  
It surpasses larger models including Idefics3~\cite{idefics}, LLaVA-Next~\cite{llava-next-interleave}, and SV-RAG~\cite{chen2024sv} by \textbf{+6.0}, \textbf{+9.5}, and \textbf{+5.2 pp}, respectively, while using \textbf{5.4$\times$} fewer visual tokens and \textbf{75\%} fewer parameters.  
These results highlight \model's strong cross-modal generalization, particularly valuable for edge devices, as it eliminates the need to load separate models for different domains.

\subsection{Ablation Studies} We report ablation on the Mp-DocVQA dataset~\cite{mpdocvqa} and LLaVA-NeXT~\cite{llava-next-interleave} as baseline model. Additional experiments are in Supplementary \Cref{sec:sup efficiency} and \Cref{sec:sup ablation}.

\begin{table}[!ht]
\centering
\scriptsize
\setlength{\tabcolsep}{4pt}
\renewcommand{\arraystretch}{1.15}
\begin{tabular}{l|ccc}
\toprule
\textbf{Modules} & \textbf{Tok/Image$\downarrow$} & \textbf{Memory(GB)$\downarrow$} & \textbf{ANLS$\uparrow$} \\
\midrule
OCR Baseline~\cite{llava-next-interleave} & 3210 & 29.2 & 50.2 \\
OCR-Free Baseline~\cite{llava-next-interleave} & 2880 & 27.9 & 38.5 \\
\hspace{2mm}(+) Visual Compression & 576 & {23.5} & 22.7 \\
\hspace{2mm}(+) OCR Compression \textcolor{gray}{(no Visual)} & 2880 & 27.9 & \underline{68.3} \\
\hspace{4mm}(+) Visual + OCR Compression & 576 & {23.5} & 67.4 \\
\rowcolor{gray!10}\hspace{6mm}(+) Streaming Abstention$^{\ast}$ & 576 & \textbf{14.2} & \textbf{70.0} \\
\bottomrule
\end{tabular}
\vspace{-1mm}
\caption{\textbf{Cumulative Ablation on Proposed Modules} on the Mp-DocVQA dataset~\cite{mpdocvqa}. Each proposed module progressively enhances text-rich visual understanding under strict token constraints. The Streaming Abstention mechanism achieves the highest accuracy while requiring the lowest GPU memory usage. $^{\ast}$Indicates the final \model~model.}
\vspace{-7mm}
\label{tab:module}
\end{table}
\noindent\textbf{Ablation on the Proposed Modules.}
In Tab. ~\ref{tab:module}, we begin with the OCR baseline, which achieves 50.2 ANLS but relies on dense tokenization. Removing OCR tokens results in a sharp performance drop to 38.5, confirming their essential role in text-grounded reasoning. Applying {Visual Compression} reduces the token count by \textbf{5.6×} (3210→576), but slightly decreases accuracy to 36.1 due to information loss from aggressive spatial downsampling. Introducing the {OCR Compression} module restores text fidelity, achieving 68.3 ANLS under a modest 2880-token budget. Furthermore, combining both OCR and visual compression achieves an optimal balance—maintaining only 576 tokens per image while preserving 67.4 ANLS. Finally, the {Streaming Abstention} module yields the best overall performance (70.0 ANLS) under the same token constraint. This progressive improvement illustrates how hierarchical compression, combined with streaming abstention, enables efficient and reliable long document understanding.
\vspace{-3mm}
\begin{figure*}[!t]
    \centering
    \includegraphics[width=1\linewidth]{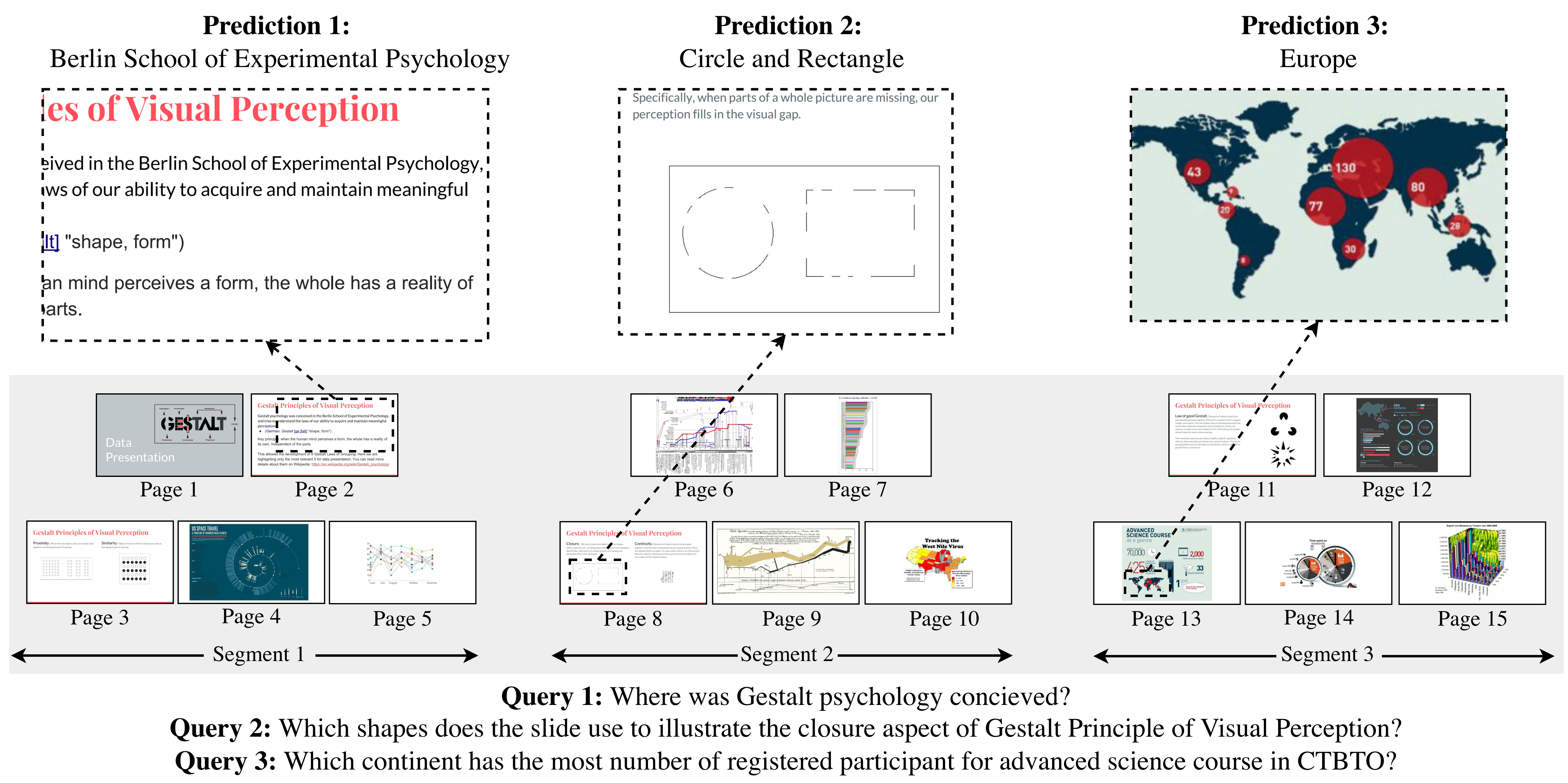}
    \caption{
    \textbf{Qualitative examples} of long-document reasoning with DocSLM.
    For each user query (bottom), DocSLM first identifies the relevant document segment via its uncertainty-based ranking mechanism before generating the final answer.  
    Queries~1–3 correspond sequentially to Segments~1–3.  
    For \textbf{Query~1}, the model reasons over text-heavy content (Page~2) to correctly identify the \textit{Berlin School of Experimental Psychology}, demonstrating effective OCR fusion.  
    For \textbf{Query~2}, it interprets visual figures to infer the shapes \textit{Circle and Rectangle}, highlighting strong visual understanding.  
    For \textbf{Query~3}, DocSLM jointly reasons over textual and visual cues in a complex map, comparing multiple numeric indicators to correctly predict \textit{Europe}, illustrating its multimodal fine-grained understanding. More results are in Supplementary \Cref{sec:sup qualitative}.
    }\vspace{-6mm}
    \label{fig:qual}
\end{figure*}

\begin{table}[!h]
\centering
\scriptsize
\setlength{\tabcolsep}{8pt}
\renewcommand{\arraystretch}{1.1}
\vspace{1mm}
\begin{tabular}{lccc}
\toprule
\textbf{Stage} & \textbf{Training Steps} & \textbf{Data Size} & \textbf{ANLS$\uparrow$} \\
\midrule
Instruction Tuning  & 3.0K & 1.00M & 38.5 \\
\hspace{1mm}(+) Image–OCR Alignment & 9.0K & 3.00M & 50.5 \\
\hspace{2mm}(+) Image Compression & 2.4K & 2.00M & 61.8 \\
\hspace{3mm}(+) Document Compression & 3.0K & 0.58M & \underline{66.7} \\
\rowcolor{gray!10}\hspace{4mm} (+)Streaming Abstention & 4.4K & 0.18M & \textbf{70.0} \\
\bottomrule
\end{tabular}
\vspace{-1mm}
\caption{\textbf{Cumulative Curriculum Training} on the Mp-DocVQA dataset~\cite{mpdocvqa}.
Each stage incrementally introduces new objectives and datasets, improving ANLS from 38.5 to 70.0.
Data size gradually decreases as task complexity increases, reflecting a shift from large-scale pretraining to specialized fine-tuning.}
\label{tab:train_stage}
\vspace{-5mm}
\end{table}

\noindent\textbf{Ablation on the Training Stages.}
Tab. \ref{tab:train_stage} demonstrates the progressive improvements achieved through our staged curriculum learning. Starting from the Instruction Tuning baseline, the model is trained without any OCR input, achieving an ANLS of 38.5\%. Introducing OCR supervision restores text grounding in the subsequent {Image–OCR Alignment} stage, with an ANLS score of 50.5\%. Adding the {Image Compression} stage greatly improves efficiency by reducing tokens by {5.6×} (3210→576) and increases ANLS to {61.8\%}. Incorporating {Document Compression} enhances long-context reasoning, achieving {66.7\%}. Finally, the proposed {Streaming Abstention} mechanism yields the best overall accuracy of {70.0\%}.

\begin{table}[!ht]
    \centering
    \scriptsize
    \setlength{\tabcolsep}{5pt}
    \renewcommand{\arraystretch}{1.05}
    \begin{tabular}{lccccc}
        \toprule
        & &\multicolumn{4}{c}{\textbf{MP-DocVQA (ANLS)$\uparrow$}} \\
        \cmidrule(lr){3-6}
        \multirow{2}{*}\textbf{Model} & {\textbf{Tok/Image}$\downarrow$} & \textbf{1} & \textbf{[2,10]} & \textbf{$>$10} & \textbf{Overall} \\
        \midrule
        Baseline     & \underline{2880}   & {75.3} & 29.7 & 0.7 & 38.5 \\
        Baseline + OCR & 3210 & \underline{78.6} & \underline{40.5} & \underline{2.2} & \underline{50.2} \\
        \rowcolor{gray!10} Ours   & \textbf{576}         & \textbf{79.6} & \textbf{70.0} & \textbf{61.2} & \textbf{70.0} \\
        \bottomrule
    \end{tabular}
    \caption{\textbf{Ablation across document lengths.}
    ANLS scores (\%) for varying numbers of pages. Our model achieves the best overall robustness, particularly on longer documents.}
    \vspace{-7mm}
    \label{tab:mpdocvqa}
\end{table}

\noindent\textbf{Ablation across Document Lengths.}
As seen in Table~\ref{tab:mpdocvqa}, the baseline model performs well on single-page inputs, but collapses on longer ones, dropping from 75.3 on single-page to just 0.7 on documents exceeding 10 pages. 
Incorporating OCR improves overall accuracy by +11.7 points, but the gain remains marginal for long documents. 
In contrast, \model~ delivers substantial improvements across all length achieving gains of +4.3, +40.3, and +60.5 ANLS for 1, [2–10], and $>$10-page documents, respectively—a far smaller degradation as the input grows. 

\noindent\textbf{Ablation on Compression Design and OCR Quality.}
(a) As shown in Table~\ref{tab:compression_ocr_ablation}, increasing the compression depth to 4 layers for both OCR and visual branches yields the best performance (70.2 ANLS), while a lightweight 2-layer configuration attains nearly identical accuracy (70.0) with 38.6M fewer parameters. We adopt this 2-layer setup as our default for all subsequent experiments.
(b) Progressive fine-tuning—first adapting linear adapters, then the SigLIP vision encoder—further enhances performance from 66.5 to 70.0 ANLS, demonstrating the benefit of joint optimization across modalities.
(c) Finally, OCR quality plays a pivotal role: PaddleOCR improves recognition accuracy to 70.0 compared to 69.1 with Tesseract, indicating that reliable text extraction remains a crucial factor specially under aggressive token compression. 
\vspace{-3mm}
\begin{table}[!t]
\centering
\scriptsize
\setlength{\tabcolsep}{4pt}
\renewcommand{\arraystretch}{1.1}
\begin{tabular}{ccc|lc|lc}
\toprule
\multicolumn{3}{c|}{\textbf{(a) Compression Depth}} &
\multicolumn{2}{c|}{\textbf{(b) Compression Tuning}} &
\multicolumn{2}{c}{\textbf{(c) OCR Source}} \\
\cmidrule(lr){1-3} \cmidrule(lr){4-5} \cmidrule(lr){6-7}
\textbf{OCR$\downarrow$} & \textbf{Visual$\downarrow$} & \textbf{ANLS$\uparrow$} &
\textbf{Tunable Params} & \textbf{ANLS$\uparrow$} &
\textbf{Source} & \textbf{Acc$\uparrow$} \\
\midrule
4 & 4 & \textbf{70.2} & Compressor & 66.5 & Tesseract & 69.1 \\
4 & \textbf{2} & 57.3 & (+) Linear Adapter & 68.6 &  PaddleOCR & \textbf{70.0} \\
\textbf{2} & 4 & 70.1 & (+) Visual Encoder & \textbf{70.0} &  &  \\
\textbf{2} & \textbf{2} & 70.0 &  &  &  &  \\
\bottomrule
\end{tabular}
\vspace{-1mm}
\caption{\textbf{Ablations on compression design and OCR quality} on the Mp-DocVQA dataset~\cite{mpdocvqa}.
(a) Balanced compression depth (4 layers each) yields the best accuracy, while a 2-layer setup achieves comparable performance with 38.6M fewer parameters. 
(b) Gradual fine-tuning of the adapter and encoder improves ANLS. 
(c) High-quality OCR provides a clear accuracy boost.}
\vspace{-7mm}
\label{tab:compression_ocr_ablation}
\end{table}

\section{Conclusion, Limitations, and Future Work}
DocSLM employs a compact architecture and feature representation for efficient, reliable multimodal understanding under strict memory constraints, enabling deployment on resource-limited devices. Although trained on limited video data, it already demonstrates strong cross-modal generalization. Future work will extend to additional modalities like audio and pursue balanced document–video training toward an omnimodal foundation model for edge deployment.

\newpage


\clearpage
\maketitlesupplementary
\setcounter{section}{0}
\setcounter{figure}{0}
\setcounter{table}{0}

\renewcommand{\thesection}{S\arabic{section}}
\renewcommand{\thetable}{T\arabic{table}}
\renewcommand{\thefigure}{F\arabic{figure}}

Our supplementary materials contain \Cref{sec:sup implementation}: Additional Implementation Details,  \Cref{sec:sup edge}: Edge Deployment, and \Cref{sec:sup efficiency}: Efficiency Analysis, \Cref{sec:sup ablation}: Additional Ablation Studies, \Cref{sec:sup qualitative}: Additional Qualitative Results.

\section{Additional Implementation Details}
\label{sec:sup implementation}

\subsection{OCR Integration}

\begin{figure}[!t]
    \centering
    \includegraphics[width=1\linewidth]{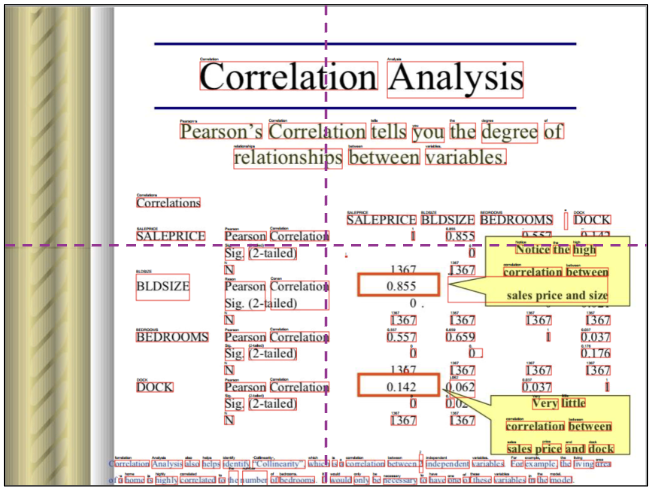}
    \caption{\textbf{OCR-to-crop assignment.}
    The OCR bounding boxes (red) are tested for overlap with the crop regions. 
    An OCR token is assigned to a crop if its bounding boxes intersect, ensuring spatially consistent OCR alignment across crops.}
    \label{fig:ocr_bbox}
\end{figure}

\begin{figure*}[!t]
    \centering
    \includegraphics[width=1\linewidth]{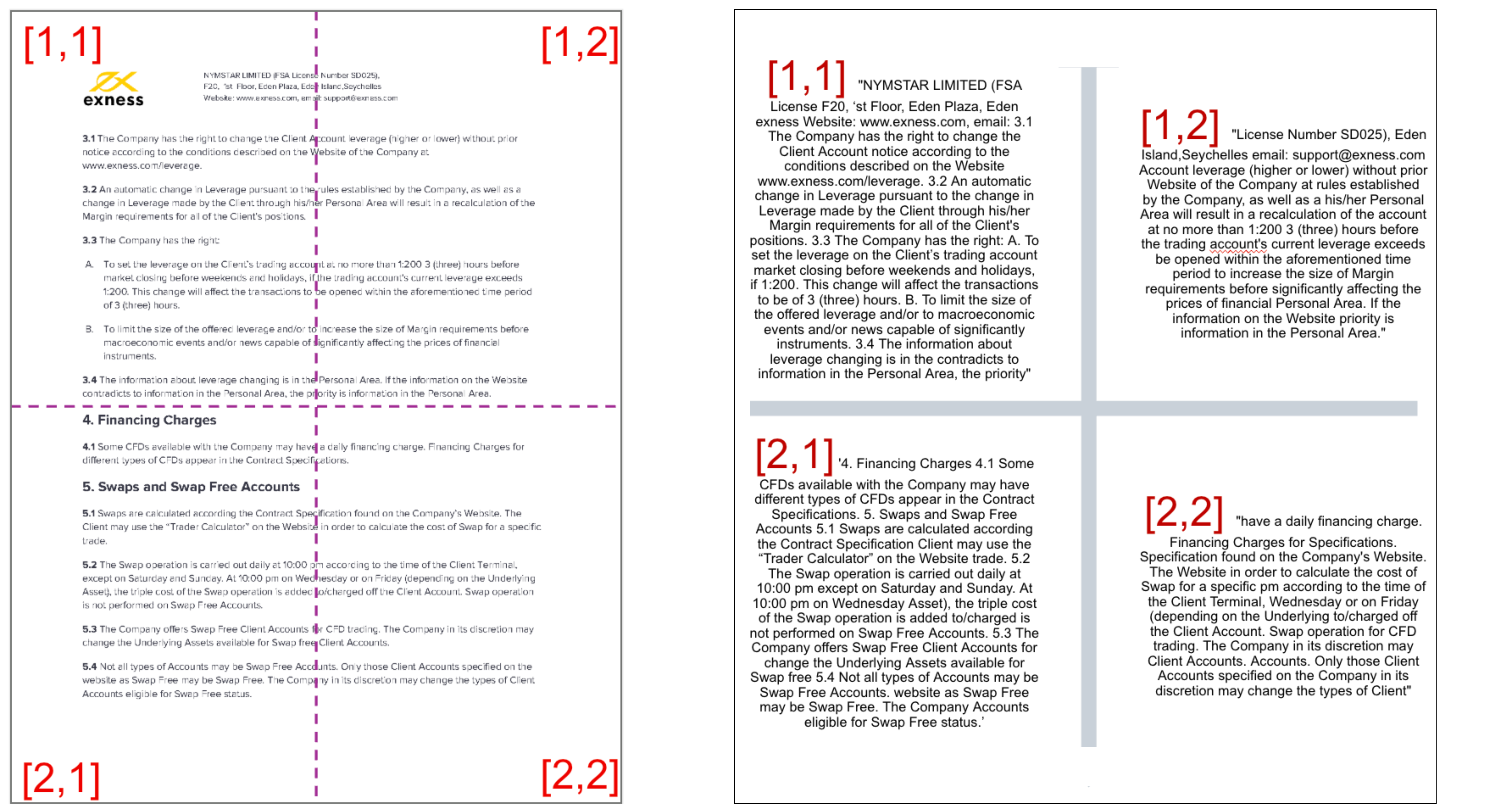}
    \caption{\textbf{Multi-crop OCR decomposition.} 
 \textbf{(left)} Each page is first resized and padded, then dynamically divided into an aspect-ratio–dependent grid of overlapping crops. \textbf{(right)} OCR tokens are spatially redistributed to their corresponding crops, enabling localized grounding and improving fine-grained multimodal alignment.}
    \label{fig:ocr_grid}
\end{figure*}

Our OCR pipeline is built around a bounding-box alignment mechanism (Fig.~\ref{fig:ocr_bbox}) that enables consistent OCR integration under multi-crop processing~\cite{llava-next-interleave} to handle documents of any size and shape. As illustrated in Fig.~\ref{fig:ocr_grid}, each input page is first resized, padded, and subdivided into a grid of non-overlapping crops. OCR tokens detected on the original image must therefore be remapped to these crops in a geometrically consistent manner. Fig. \ref{fig:ocr_bbox} and \ref{fig:ocr_grid} show 4 regions for simplicity; however, based on aspect ratio and resolution, the number of crops can range from 4-16 in our default setup. As visualized in Fig.~\ref{fig:ocr_bbox}, each OCR token is represented by a bounding box in original-image coordinates, normalized by the image width and height. A token is assigned to every crop whose bounding box overlaps with it. This supports one-to-many assignments when a word spans crop boundaries and handles empty crops. The resulting crop-aligned OCR lists are then fused with the hierarchical multimodal compression module. This alignment mechanism ensures that multimodal training receives consistent and spatially grounded OCR information, even under highly variable document layouts and multi-resolution patch configurations.

\subsection{Training Details}
Table \ref{tab:train_config_scale} summarizes the full five-stage training pipeline used to build our 2B-parameter model. The training strategy gradually transitions from large-scale noisy pretraining to highly curated downstream finetuning, while progressively increasing task difficulty and reducing learning rates. Pretrain 1 initializes the multimodal alignment by training the MLP adapter and hierarchical compressor on 1M weakly supervised image–text pairs using cross-attention–based fusion of SigLIP2\cite{tschannen2025siglip} visual features and PaddleOCR\cite{li2022paddleocr} tokens. Pretrain 2 scales the same objective to a larger 3M corpus and unlocks the vision tower and multimodal compressor for joint optimization, improving cross-modal grounding. Pretrain 3 adapts the model to high-quality single-page document datasets (2M samples), introduces early-layer OCR and visual compression, and begins tuning the language model to better handle structured document semantics. Finetune 1 transitions to the DocDownstream-1.0\cite{hu2024mplug_docowl_1_5} mixture (0.58M examples) and trains under long-context settings, enabling robust reasoning over long documents while maintaining a manageable batch size via ZeRO-2 and gradient accumulation\cite{feng2021optimal}. Finally, Finetune 2 introduces negative-pair supervision and multi-image document sequences, training the model to abstain on unsupported evidence and improving calibration in streaming settings. 

\begin{table}[!t]
    \centering
    \scriptsize
    \setlength{\tabcolsep}{8pt}
    \renewcommand{\arraystretch}{1.25}
    \vspace{1mm}
    \begin{tabular}{lcccc}
        \toprule
        \textbf{Stage} & \textbf{Training Steps} & \textbf{Batch} & \textbf{Data Size} & \textbf{LR} \\
        \midrule
        Pretrain~1 & 3.0K & 1.0K & 1.00M & $1\!\times\!10^{-4}$ \\
        Pretrain~2 & 9.0K & 1.0K & 3.00M & $1\!\times\!10^{-4}$ \\
        Pretrain~3 & 2.4K & 1.0K & 2.00M & $2\!\times\!10^{-5}$ \\
        Finetune~1 & 3.0K & 256  & 0.58M & $2\!\times\!10^{-5}$ \\
        \rowcolor{gray!8}Finetune~2 & 4.4K & 256  & 0.18M & $2\!\times\!10^{-6}$ \\
        \bottomrule
    \end{tabular}
    \vspace{-1mm}
    \caption{Each stage progressively adapts to more complex tasks, while the availability of high-quality data decreases.}
    \label{tab:train_config_scale}
    \vspace{-2mm}
\end{table}

\noindent Across all stages, we use bf16 precision and flash-attention \cite{dao2022flashattention}. This staged progression allows the model to retain broad generalization from large-scale pretraining while acquiring strong long-document reasoning capabilities from high-quality downstream data. To further boost computational and memory efficiency, we incorporate Liger Kernel~\cite{dai2024ligerkernel}, a lightweight optimization toolkit designed for large-scale model training. Liger provides high-performance fused operators and memory-aware execution strategies, such as combining sequential kernels, using in-place updates, and partitioning inputs into manageable chunks. These optimizations increase training throughput while lowering the memory footprint, enabling our multimodal model to scale more effectively under constrained GPU resources. The complete implementation details can be found in the attached codebase.

\section{Edge Deployment}
\label{sec:sup edge}
\begin{figure*}[!t]
    \centering
    \includegraphics[width=1\linewidth]{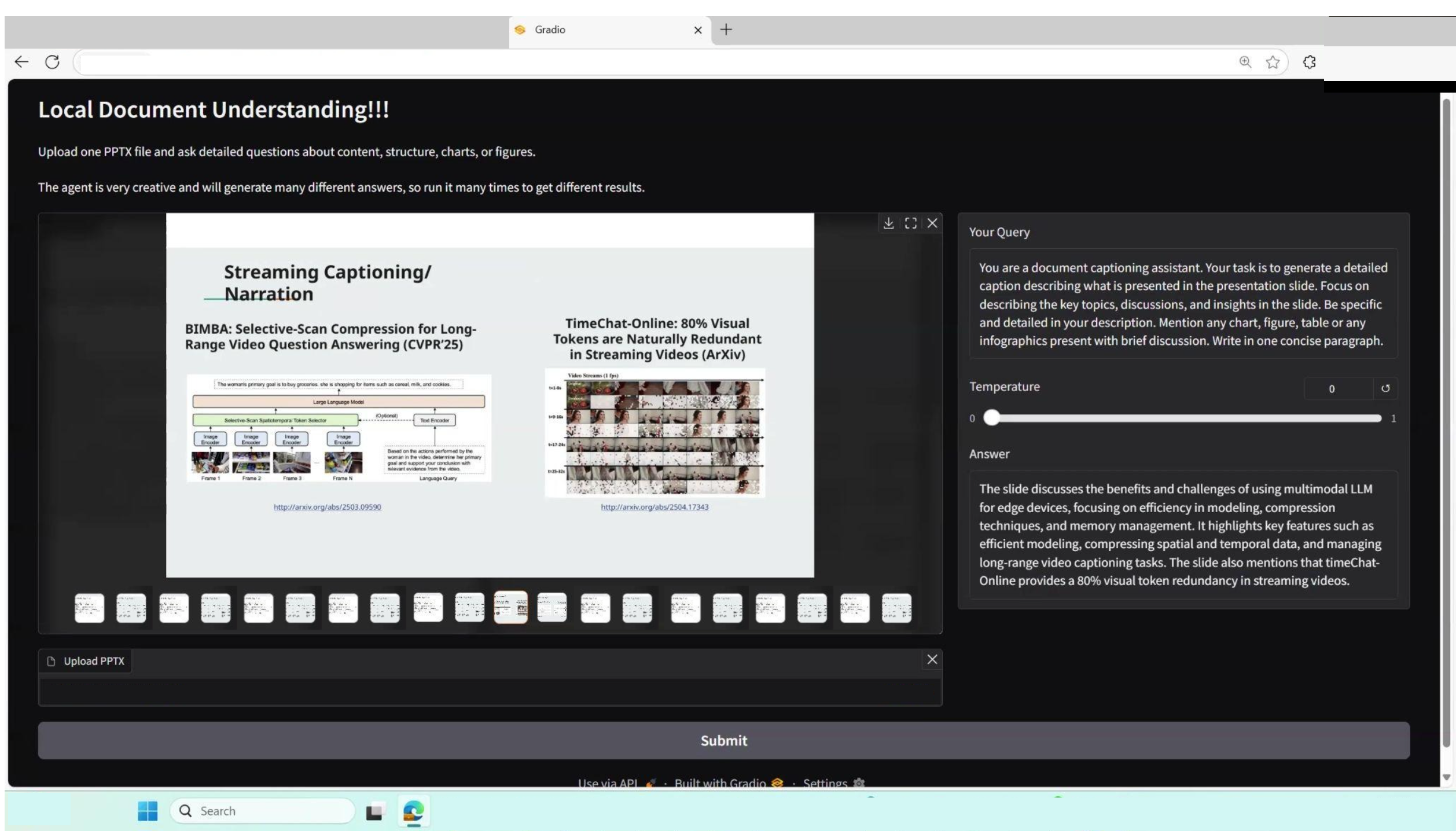}
    \caption{\textbf{Local Document Understanding on Laptop.} 
    Screenshot of our interactive on-device system for local document understanding. Users can upload PPTX files, browse slide thumbnails, and issue natural-language queries about slide content, structure, or figures. Responses are generated entirely on-device using a Windows laptop powered by a Qualcomm Snapdragon X Elite (X1E80100) with 16 GB memory. This setup demonstrates that our pipeline performs fine-grained multimodal reasoning locally on lightweight edge hardware without relying on cloud resources. Portions of the interface have been \textbf{anonymized} using solid color blocks.}
    \label{fig:edge_interface}
\end{figure*}

To enable fast and memory-efficient on-device inference, we convert our PyTorch-based Vision-Language Model into an optimized NPU-executable pipeline through a sequence of conversion and hardware-specific compilation steps to run on a Windows Copilot+ Laptop\cite{Buy138in63:online} (Fig. \ref{fig:edge_interface}).

\vspace{0.5em}
\noindent\textbf{1. ONNX~\cite{Executio91:online} Conversion.}
The PyTorch model is first exported to the {ONNX} format using the standard PyTorch tracing pipeline.  
The exported ONNX graph preserves full model parameters, operator structure, and tensor formats required for downstream compiler optimization.  
This intermediate representation provides a hardware-agnostic bridge between the PyTorch runtime and the target NPU execution environment.

\vspace{0.5em}
\noindent\textbf{2. Weight and Activation Quantization.}
We then apply post-training quantization to the full ONNX model.  
All model weights are quantized to \textbf{8-bit integers} using a min--max calibration scheme, while activations are quantized to \textbf{16-bit} precision.  
Quantization statistics, the scales and offsets of the layers are computed from a representative set of \textbf{300 document samples}.  
The resulting quantized model runs natively on both GPU and CPU backends in PyTorch, enabling thorough validation before hardware conversion.

\vspace{0.5em}
\noindent\textbf{3. NPU Compilation.}
Finally, we compile the Quantized ONNX model using the {Qualcomm AI Engine (QNN) compiler}~\cite{Qualcomm95:online} to generate a fully NPU-executable binary.  
The compiler maps ONNX operators to NPU-supported kernels, performs graph-level optimizations, and produces a hardware-targeted model artifact.  
This step transforms the architecture into a latency-optimized, memory-efficient {NPU-runnable} Vision-Language model while retaining the core multimodal reasoning capabilities of the original implementation. Specifically, we use a Windows laptop equipped with a Snapdragon X Elite (X1E80100) processor featuring a 45-TOPS Hexagon-class NPU and 16 GB of unified memory.

\noindent This deployment pipeline enables our model to run efficiently on edge devices, substantially reducing memory consumption while sustaining high throughput.  
It provides a practical path for real-world applications such as slide analysis, document assistants, and on-device multimodal agents.

\section{Additional Efficiency Analysis}
\label{sec:sup efficiency}

\paragraph{Memory Efficiency.}
Following standard memory analyses of transformer architectures~\cite{dao2022flashattention,dao2023flashattention,dettmers2023qlora}, 
the peak VRAM during inference can be expressed using the simple approximation:
\begin{equation}
\label{eq:mem}
\text{VRAM}_{\text{peak}} \;\approx\;
\underbrace{P_{\text{B}} \times b}_{\substack{\text{parameter}\\\text{memory}}}
\;+\;
\underbrace{K \times g}_{\substack{\text{KV-cache}\\\text{(per 1k tokens)}}}
\;+\;
\underbrace{\mathcal{O}}_{\substack{\text{fixed}\\\text{overhead}}}
\end{equation}
where $P_{\text{B}}$ is the number of parameters (in billions), $b$ is the bytes per parameter, 
$K$ is the context length measured in units of $1$k tokens, $g$ is the KV-cache cost per $1$k tokens, 
and $\mathcal{O}$ denotes fixed activation and workspace memory.  
Existing document VLMs typically emit $3$k--$4$k visual tokens per page, which leads to a steep linear increase in the 
token-dependent term $K g$ as the number of pages grows. 
Our model follows the same linear trend in principle; however, the crucial difference is the \emph{slope} of this growth. 

\begin{figure}[!t]
    \centering
    \includegraphics[width=1\linewidth]{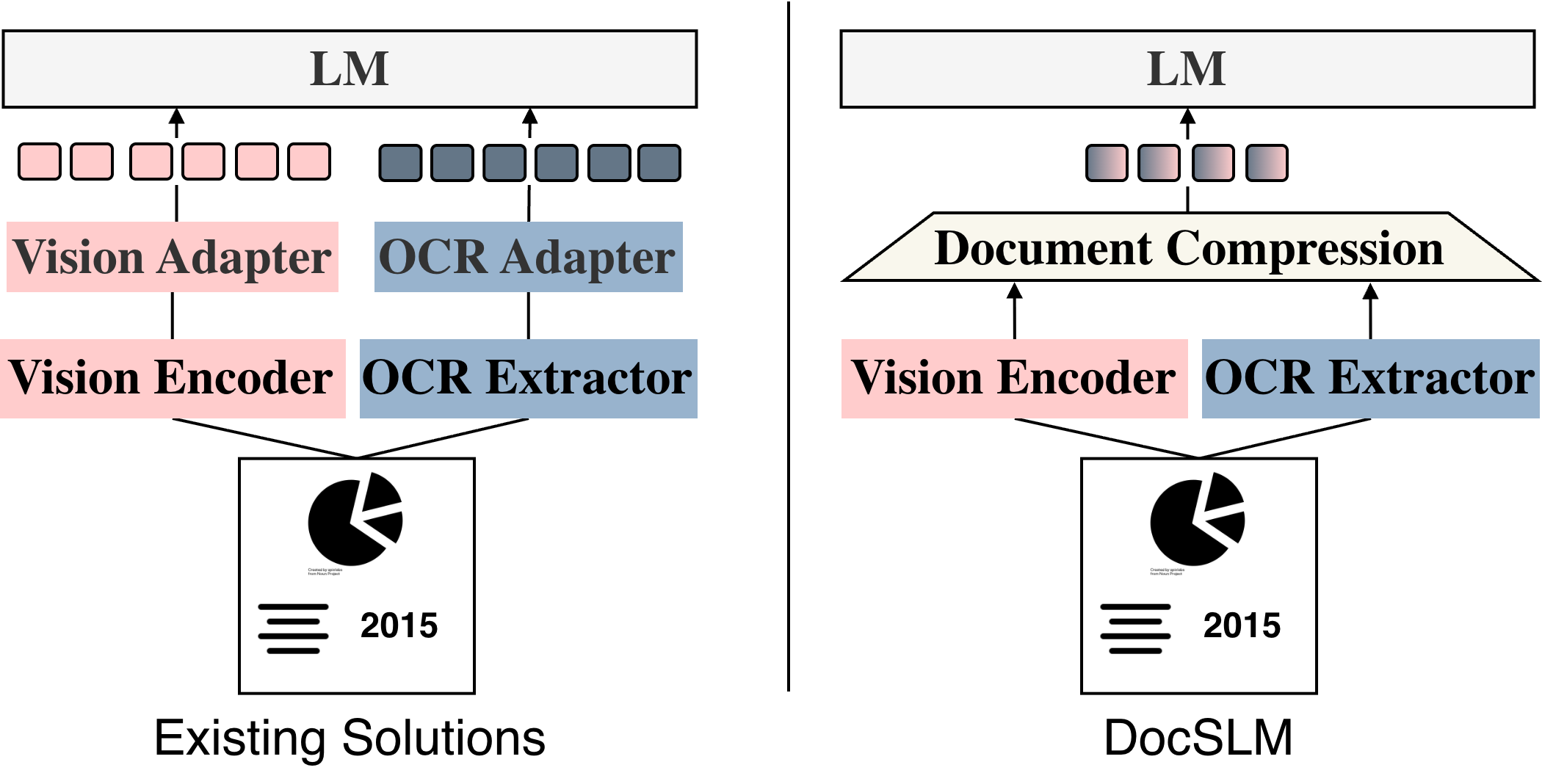}
    \caption{
    Prior methods process visual and OCR features independently, resulting in a large number of input tokens for the language model. 
    In contrast, \model~ fuses both modalities with a compression module, substantially reducing token count.
    }
    \label{fig:compare}
\end{figure}

\noindent DocSLM compresses OCR, visual, and layout information into a fixed \textbf{576-token} representation per page, which dramatically 
reduces $K$ for any given document.  
As a result, the contribution of the KV-cache and activation components in Eq.~\ref{eq:mem} grows much more slowly for our model, 
yielding a significantly lower overall memory footprint across long documents compared to baselines whose vision encoders 
produce thousands of tokens or crops per page.

\paragraph{Peak GPU Memory Comparison}

Table~\ref{tab:memory_page} reports the peak GPU memory usage of several Document understanding models as the number of document pages increases from 2 to 120. All experiments were conducted using the official implementations of each model on an NVIDIA A100–80GB GPU, using the MMLongDocBench~\cite{ma2024mmlong} dataset. We observe that existing large and medium-scale models (InternVL2-RAG\cite{wang2024needle}, Docopilot\cite{duan2025docopilot}, DocOWL2\cite{ye2023mplugdocowl}) exhibit monotonic memory growth as document length increases, with memory rising sharply between 10 and 20 pages before eventually triggering out-of-memory failures. This behavior highlights the fundamental limitation of these architectures (Fig. \ref{fig:compare}) whose token counts scale linearly with the number of pages. In contrast, our streaming 2B model maintains a strictly constant peak memory footprint of 14.2 GB across all document lengths—including the 120-page setting—due to its fixed-size per-page multimodal representation and sequential stream processing. This plateau demonstrates that our design fully decouples memory usage from document length, enabling reliable, large-scale document understanding on fixed-memory hardware such as edge GPUs, laptops, and resource-constrained servers.

\begin{table}[!h]
\centering
\scriptsize
\setlength{\tabcolsep}{6pt}
\renewcommand{\arraystretch}{1.15}

\begin{tabular}{l c  ccccccc}
\toprule
\multirow{2}{*}{\textbf{Model}}
& \multirow{2}{*}{\textbf{Size}}
& \multicolumn{6}{c}{\textbf{Peak Memory (GB) by Page Count}} \\
\cmidrule(lr){3-8}
&  & \textbf{2} & \textbf{5} & \textbf{10} & \textbf{15} & \textbf{20} & \textbf{120} \\
\midrule

InternVL2-RAG~\cite{wang2024needle}
& 8B & 22.6 & 31.7 & 47.0 & 61.9 & 76.8 & \texttt{OOM} \\

Docopilot\cite{duan2025docopilot}
& 8B & 21.6 & 30.5 & 45.5 & 60.4 & 75.3 & \texttt{OOM} \\

InternVL2-RAG~\cite{wang2024needle}
& \textbf{2B} & 10.8 & 18.2 & 30.3 & 42.9 & 55.3 & \texttt{OOM} \\

Docopilot~\cite{duan2025docopilot}
& \textbf{2B} & \underline{9.2} & \underline{16.2} & 27.9 & 40.3 & 52.7 & \texttt{OOM} \\

DocOWL2~\cite{hu2024mplugdocowl2}
& 8B & 17.7 & 20.0 & \underline{24.4} & \underline{28.6} & \underline{34.1} & \texttt{OOM} \\

\rowcolor{gray!10}
\textbf{Ours}
& \textbf{2B} & \textbf{5.2} & \textbf{9.2} & \textbf{14.2} & \textbf{14.2} & \textbf{14.2} & \textbf{14.2} \\

\bottomrule
\end{tabular}

\caption{
\textbf{Peak GPU memory usage (GB) under increasing document length.}
Measurements were obtained on an NVIDIA A100--80GB GPU using the
MMLongDocBench~\cite{ma2024mmlong} dataset. Our streaming 2B model
maintains a constant 14.2\,GB memory footprint up to 120 pages.
}
\label{tab:memory_page}
\end{table}

\paragraph{Latency Vs. Accuracy} 
Table \ref{tab:mmlongdoc_latency} presents a detailed comparison of inference latency and accuracy on the MMLongDoc\cite{ma2024mmlong} benchmark across a range of state-of-the-art large multimodal models. Existing LVLMs, such as InternVL2-RAG\cite{wang2024needle} (2B/8B), Docopilot-2B\cite{duan2025docopilot}, and VisRAG-12B\cite{yu2024visrag} exhibit high computational overhead due to their large parameter counts and heavy visual token budgets (approximately 3K tokens per image). Even with retrieval-augmented pipelines (InternVL2+RAG), latency remains high (82–113 ms) and accuracy does not improve, highlighting the limitations of RAG-based pruning for long-document reasoning.

In contrast, our 2B model uses only 576 tokens per image through hierarchical multimodal compression, resulting in a 3–7× reduction in latency while simultaneously achieving the highest accuracy (22.7 Acc). This efficiency–accuracy trade-off demonstrates that compact models, when paired with structured compression and streaming mechanisms, can outperform much larger LVLMs both in speed and effectiveness, making our approach particularly suitable for real-time and edge-device deployment.

\begin{table}[!h]
\centering
\setlength{\tabcolsep}{6pt}
\renewcommand{\arraystretch}{1.1}
\scriptsize
\begin{tabular}{lcccc}
\toprule
\textbf{Model} & \textbf{Size} & \textbf{Tok/Image↓} & \textbf{Latency (ms)↓} & \textbf{MMLDoc (Acc↑)} \\
\midrule
InternVL2           & 8B   & $\sim$3,133   & 81.0  & 17.4 \\
InternVL2+RAG       & 2B   & $\sim$3,133   & 82.9  & 17.2 \\
VisRAG       & 12B   & $>$3K   & 288.3  & 18.8 \\
InternVL2             & 2B   & $\sim$3,133   & 35.9  & 10.5 \\
Docopilot             & 2B   & $\sim$3,133   & 35.9  & 21.8 \\
\rowcolor{gray!10}
\textbf{Ours}       & \textbf{2B} & \textbf{576} & \textbf{32.1} & \textbf{22.7} \\
\bottomrule
\end{tabular}
\caption{
\textbf{Latency vs. accuracy} comparison on {MMLongDoc~\cite{ma2024mmlong} (Acc)}. Our 2B model achieves SOTA accuracy with substantially lower latency.
}\vspace{-5mm}
\label{tab:mmlongdoc_latency}
\end{table}

\section{Additional Ablation Studies}
\label{sec:sup ablation}
\subsection{Effect of OCR Confidence Threshold}

To evaluate our model's robustness to OCR noise, we apply a confidence filter to OCR tokens before fusion:
\begin{equation}
\mathcal{T}_{\text{OCR}}(\tau)
= 
\left\{\, t \in \mathcal{T} \;\middle|\; \mathrm{conf}(t) \ge \tau \right\},
\label{eq:ocr_threshold}
\end{equation}

\noindent where $\tau$ is the OCR confidence threshold.  
Table~\ref{tab:ocr_threshold_scaled} reports Mp-DocVQA accuracy for thresholds ranging from 0.0 to 0.9. Performance remains extremely stable across the full range, with the best result at $\tau=0.0$. This indicates that our hierarchical compressor effectively absorbs OCR noise, and that aggressive filtering may remove useful but low-confidence text tokens.

\begin{table}[!h]
\centering
\small
\begin{tabular}{lcccccc}
\toprule
\textbf{OCR Threshold} & 0.0 & 0.5 & 0.6 & 0.7 & 0.8 & 0.9 \\
\midrule
\textbf{Mp-DocVQA}     & \textbf{70.0} & 69.6 & 69.6 & 69.7 & 69.7 & 69.4 \\
\bottomrule
\end{tabular}
\caption{\textbf{Ablation on OCR confidence threshold.}  
Performance remains consistent across all thresholds, indicating that our model is robust to OCR noise and does not rely heavily on aggressive confidence filtering.}
\label{tab:ocr_threshold_scaled}
\end{table}

\begin{figure*}[!t]
    \centering
    \includegraphics[width=1\linewidth]{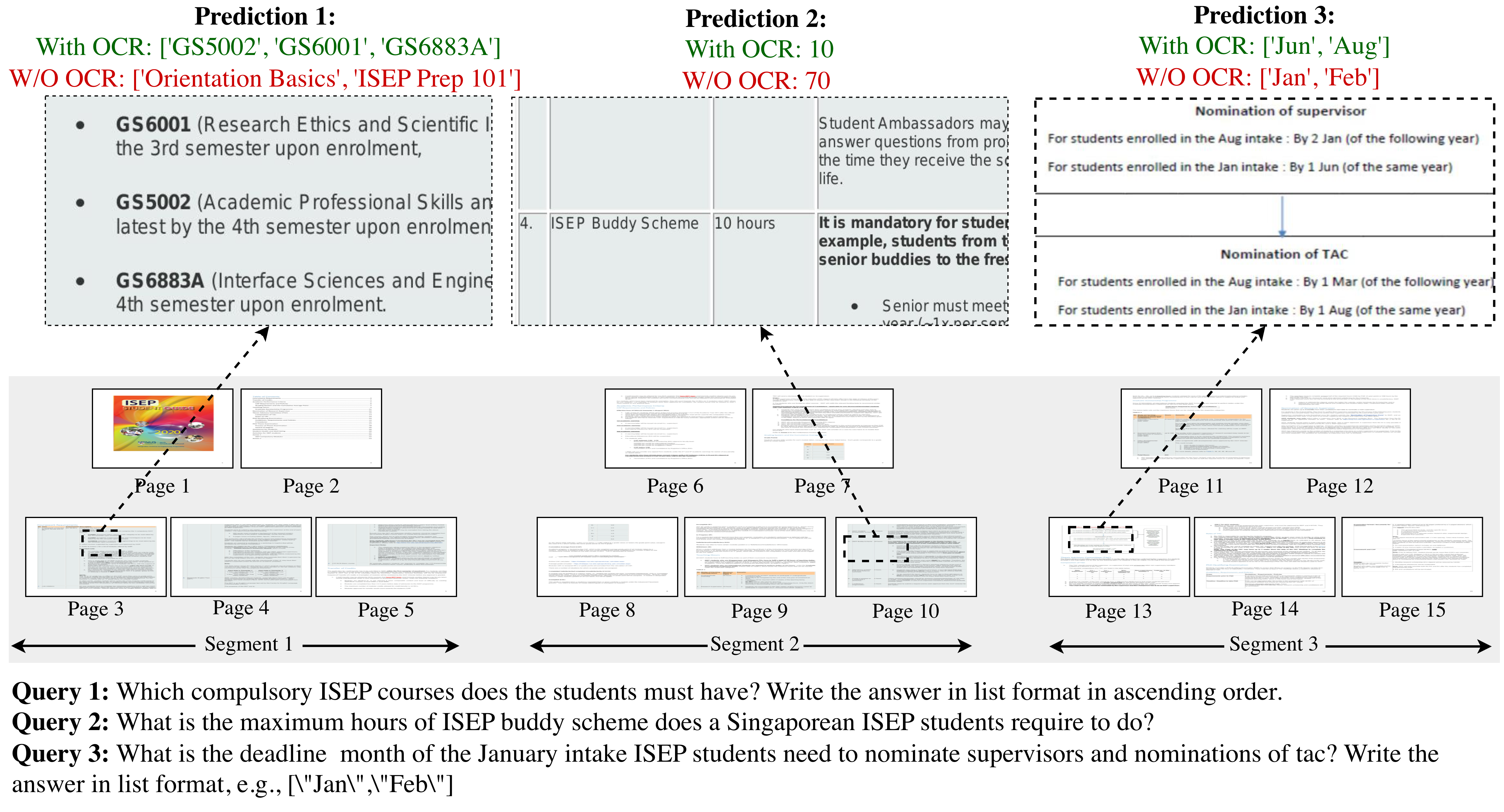}
    \caption{\textbf{Qualitative comparison of model predictions with and without OCR on a 15-page text-rich document.} With OCR (green), the model extracts the correct answers directly from the corresponding pages (highlighted). Without OCR (red), the model fails to recognize text-dense regions, instead hallucinating plausible-sounding but incorrect outputs. This illustrates that the failure arises from missing text perception rather than reasoning when processing visually complex document layouts.}
    \label{fig:qual_ocr}
\end{figure*}

\subsection{Ablation: Effect of OCR Granularity}
\label{sec:sup_ocr_granularity}
To study how OCR granularity influences model performance, we evaluate three configurations of the dynamic cropping pipeline, each corresponding directly to one entry in Table~\ref{tab:ocr_granularity}.
Specifically:
(i) fine-grained cropping (768–2304), which produces the largest number of crops (16–100),
(ii) medium cropping (384–1536), which generates a moderate number of crops (4–49), and
(iii) coarse cropping (384–1152), which yields the smallest crop count (4–18).
These settings differ in the density of visual patches produced by the dynamic cropping pipeline and, accordingly, the locality of OCR tokens grounded within each patch. Table~\ref{tab:ocr_granularity} reports MP-DocVQA accuracy for all three configurations.
The coarse configuration (384–1152) achieves the highest accuracy, while both the medium and especially the fine-grained configurations underperform despite introducing more crops and enabling more localized OCR grounding. Higher-resolution cropping grids (e.g., 768–2304) fragment each page into many small overlapping patches, forcing OCR tokens to be split across numerous local regions.

\begin{table}[!h]
\centering
\scriptsize
\setlength{\tabcolsep}{10pt}
\renewcommand{\arraystretch}{1.15}

\begin{tabular}{c c c}
\toprule
\textbf{Resolution} & \textbf{\#Crops} & \textbf{MP-DocVQA} \\
\midrule
768--2304  & 16--100 & 56.7 \\
384--1536  & 4--49   & 57.9 \\
\rowcolor{gray!10} 384--1152  & 4--18   & \textbf{70.0} \\
\bottomrule
\end{tabular}

\caption{\textbf{Ablation on OCR granularity across dynamic crop configurations.}
The \#Crops column indicates the range of possible crops generated for each resized resolution; the exact number depends on the aspect ratio of the original document.  
Mid-range resolutions (384--1152) achieve the best balance between OCR locality and global structure.}
\label{tab:ocr_granularity}
\end{table}

Although this improves fine-grained text–vision alignment, it disrupts global document structure, paragraph continuity, table layout, and multi-column flow—which hinders holistic document understanding.
As a result, finer-grained OCR assignments do not yield performance gains and instead degrade accuracy. In contrast, the coarse configuration (384–1152) preserves global layout while still providing adequate OCR grounding for local reasoning. This balance enables the hierarchical compressor to integrate textual cues without over-fragmenting the document. Overall, these results show that higher OCR granularity does not necessarily improve performance.
Effective long-document understanding requires a balance between local OCR grounding and global structural coherence, and the coarse 384–1152 configuration offers the most favorable trade-off.

\begin{figure*}[!t]
    \centering
    \includegraphics[width=1\linewidth]{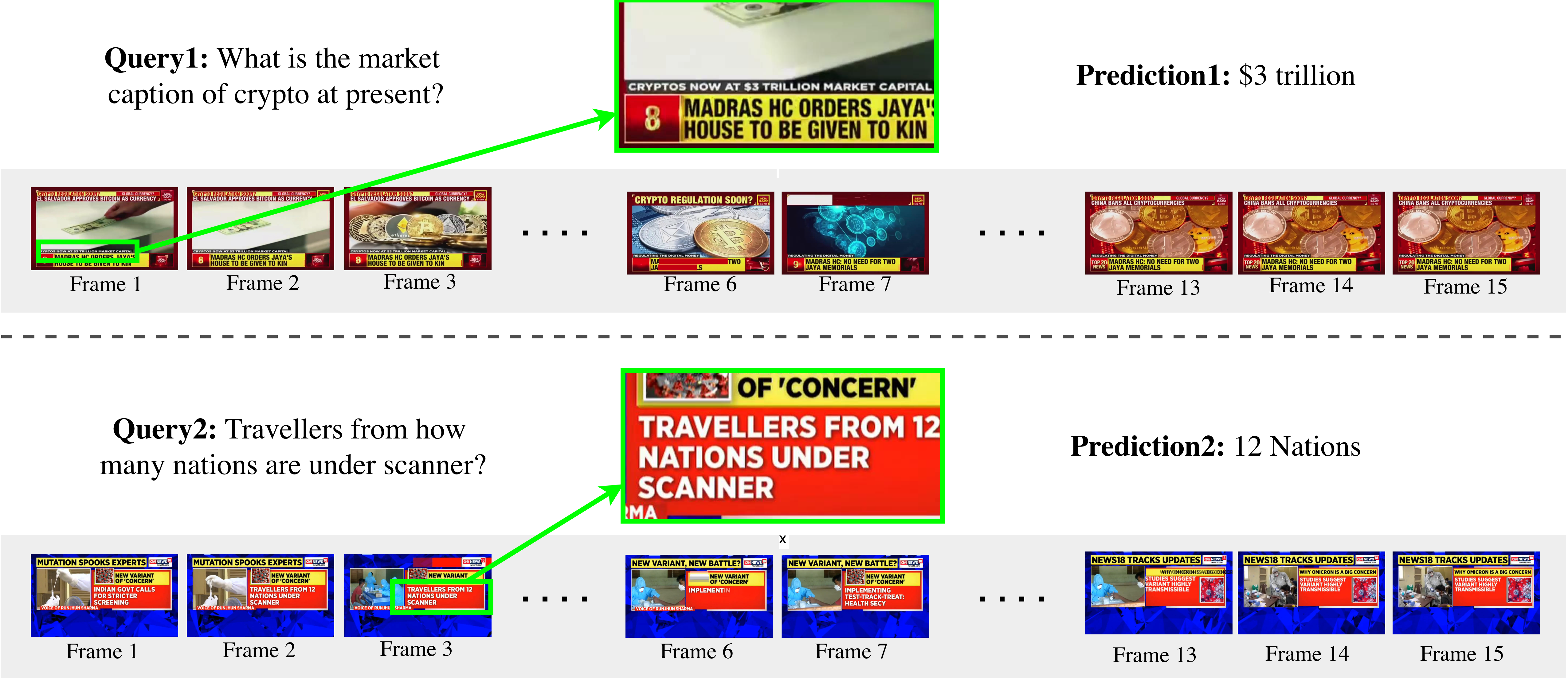}
    \caption{\textbf{Qualitative examples of generalization to videos.} We evaluate our model on the NewsVQA \cite{newsvideoqa} benchmark, which requires understanding text embedded within video frames. We show two representative cases where our model accurately identifies the temporal segment containing the answer and correctly interprets the textual cues present in the frames. These examples highlight the model’s ability to leverage multimodal signals for precise temporal localization and factually grounded answering in real video scenarios.}
    \label{fig:qual_newsvqa}
\end{figure*}

\section{Aditional Qualitative Results}
\label{sec:sup qualitative}
\paragraph{With vs without OCR}

Fig. \ref{fig:qual_ocr} presents a qualitative analysis of the model’s behavior on a multi-page, text-heavy academic document when OCR is present versus absent. With OCR, the model consistently retrieves correct information from the relevant pages, demonstrating reliable grounding across segments (Pages 3, 10, and 13). In contrast, without OCR the model is unable to parse dense textual regions and instead hallucinates answers that bear no relation to the document content (e.g., inventing course names, misreading table quantities, and guessing arbitrary deadline months). These errors highlight a fundamental limitation of vision-only processing: the model fails not due to reasoning but due to its inability to perceive fine-grained text embedded in complex layouts. This underscores the necessity of OCR for long-document understanding tasks requiring precise textual extraction.

\paragraph{Generalization to Videos}

Fig. \ref{fig:qual_newsvqa} presents qualitative examples illustrating our model’s ability to generalize to real-world video settings. Using the NewsVQA \cite{newsvideoqa} benchmark, which demands a precise understanding of text appearing within broadcast news footage, our method successfully identifies the temporal window in which the answer-relevant information is displayed. In both cases, the model tracks the textual overlays across frames, correctly localizes the segment containing the key evidence, and produces a factually accurate answer. These results demonstrate that our approach effectively leverages fine-grained textual cues in videos, enabling robust temporal grounding and reliable question answering in dynamic, text-rich video environments.

\newpage

{
    \small
    \bibliographystyle{ieeenat_fullname}
    \bibliography{main}

@String(ICCV= {Int. Conf. Comput. Vis.})

@String(ICCV  = {ICCV})

@article{tschannen2025siglip,
  title={Siglip 2: Multilingual vision-language encoders with improved semantic understanding, localization, and dense features},
  author={Tschannen, Michael and Gritsenko, Alexey and Wang, Xiao and Naeem, Muhammad Ferjad and Alabdulmohsin, Ibrahim and Parthasarathy, Nikhil and Evans, Talfan and Beyer, Lucas and Xia, Ye and Mustafa, Basil and others},
  journal={arXiv preprint arXiv:2502.14786},
  year={2025}
}

@misc{qwen2025qwen25technicalreport,
      title={Qwen2.5 Technical Report}, 
      author={Qwen and : and An Yang and Baosong Yang and Beichen Zhang and Binyuan Hui and Bo Zheng and Bowen Yu and Chengyuan Li and Dayiheng Liu and Fei Huang and Haoran Wei and Huan Lin and Jian Yang and Jianhong Tu and Jianwei Zhang and Jianxin Yang and Jiaxi Yang and Jingren Zhou and Junyang Lin and Kai Dang and Keming Lu and Keqin Bao and Kexin Yang and Le Yu and Mei Li and Mingfeng Xue and Pei Zhang and Qin Zhu and Rui Men and Runji Lin and Tianhao Li and Tianyi Tang and Tingyu Xia and Xingzhang Ren and Xuancheng Ren and Yang Fan and Yang Su and Yichang Zhang and Yu Wan and Yuqiong Liu and Zeyu Cui and Zhenru Zhang and Zihan Qiu},
      year={2025},
      eprint={2412.15115},
      archivePrefix={arXiv},
      primaryClass={cs.CL},
      url={https://arxiv.org/abs/2412.15115}, 
}

@article{guan2025token,
  title={A token-level text image foundation model for document understanding},
  author={Guan, Tongkun and Wang, Zining and Fu, Pei and Guo, Zhengtao and Shen, Wei and Zhou, Kai and Yue, Tiezhu and Duan, Chen and Sun, Hao and Jiang, Qianyi and others},
  journal={arXiv preprint arXiv:2503.02304},
  year={2025}
}

@misc{yu2025docthinkerexplainablemultimodallarge,
      title={DocThinker: Explainable Multimodal Large Language Models with Rule-based Reinforcement Learning for Document Understanding}, 
      author={Wenwen Yu and Zhibo Yang and Yuliang Liu and Xiang Bai},
      year={2025},
      eprint={2508.08589},
      archivePrefix={arXiv},
      primaryClass={cs.CV},
      url={https://arxiv.org/abs/2508.08589}, 
}

@misc{nacson2024docvlmmakevlmefficient,
      title={DocVLM: Make Your VLM an Efficient Reader}, 
      author={Mor Shpigel Nacson and Aviad Aberdam and Roy Ganz and Elad Ben Avraham and Alona Golts and Yair Kittenplon and Shai Mazor and Ron Litman},
      year={2024},
      eprint={2412.08746},
      archivePrefix={arXiv},
      primaryClass={cs.CV},
      url={https://arxiv.org/abs/2412.08746}, 
}

@misc{zhu2025simpleeffectivelayouttoken,
      title={A Simple yet Effective Layout Token in Large Language Models for Document Understanding}, 
      author={Zhaoqing Zhu and Chuwei Luo and Zirui Shao and Feiyu Gao and Hangdi Xing and Qi Zheng and Ji Zhang},
      year={2025},
      eprint={2503.18434},
      archivePrefix={arXiv},
      primaryClass={cs.CV},
      url={https://arxiv.org/abs/2503.18434}, 
}

@misc{xiao2025adaptivemarkuplanguagegeneration,
      title={Adaptive Markup Language Generation for Contextually-Grounded Visual Document Understanding}, 
      author={Han Xiao and Yina Xie and Guanxin Tan and Yinghao Chen and Rui Hu and Ke Wang and Aojun Zhou and Hao Li and Hao Shao and Xudong Lu and Peng Gao and Yafei Wen and Xiaoxin Chen and Shuai Ren and Hongsheng Li},
      year={2025},
      eprint={2505.05446},
      archivePrefix={arXiv},
      primaryClass={cs.CV},
      url={https://arxiv.org/abs/2505.05446}, 
}

@misc{liao2025doclayllmefficientmultimodalextension,
      title={DocLayLLM: An Efficient Multi-modal Extension of Large Language Models for Text-rich Document Understanding}, 
      author={Wenhui Liao and Jiapeng Wang and Hongliang Li and Chengyu Wang and Jun Huang and Lianwen Jin},
      year={2025},
      eprint={2408.15045},
      archivePrefix={arXiv},
      primaryClass={cs.CV},
      url={https://arxiv.org/abs/2408.15045}, 
}

@misc{duan2025docopilotimprovingmultimodalmodels,
      title={Docopilot: Improving Multimodal Models for Document-Level Understanding}, 
      author={Yuchen Duan and Zhe Chen and Yusong Hu and Weiyun Wang and Shenglong Ye and Botian Shi and Lewei Lu and Qibin Hou and Tong Lu and Hongsheng Li and Jifeng Dai and Wenhai Wang},
      year={2025},
      eprint={2507.14675},
      archivePrefix={arXiv},
      primaryClass={cs.CV},
      url={https://arxiv.org/abs/2507.14675}, 
}

@misc{wang2025martenvisualquestionanswering,
      title={Marten: Visual Question Answering with Mask Generation for Multi-modal Document Understanding}, 
      author={Zining Wang and Tongkun Guan and Pei Fu and Chen Duan and Qianyi Jiang and Zhentao Guo and Shan Guo and Junfeng Luo and Wei Shen and Xiaokang Yang},
      year={2025},
      eprint={2503.14140},
      archivePrefix={arXiv},
      primaryClass={cs.CV},
      url={https://arxiv.org/abs/2503.14140}, 
}

@misc{tanaka2025vdocragretrievalaugmentedgenerationvisuallyrich,
      title={VDocRAG: Retrieval-Augmented Generation over Visually-Rich Documents}, 
      author={Ryota Tanaka and Taichi Iki and Taku Hasegawa and Kyosuke Nishida and Kuniko Saito and Jun Suzuki},
      year={2025},
      eprint={2504.09795},
      archivePrefix={arXiv},
      primaryClass={cs.CL},
      url={https://arxiv.org/abs/2504.09795}, 
}

@article{chen2024sv,
  title={SV-RAG: LoRA-Contextualizing Adaptation of MLLMs for Long Document Understanding},
  author={Chen, Jian and Zhang, Ruiyi and Zhou, Yufan and Yu, Tong and Dernoncourt, Franck and Gu, Jiuxiang and Rossi, Ryan A and Chen, Changyou and Sun, Tong},
  journal={arXiv preprint arXiv:2411.01106},
  year={2024}
}

@article{yu2024visrag,
  title={Visrag: Vision-based retrieval-augmented generation on multi-modality documents},
  author={Yu, Shi and Tang, Chaoyue and Xu, Bokai and Cui, Junbo and Ran, Junhao and Yan, Yukun and Liu, Zhenghao and Wang, Shuo and Han, Xu and Liu, Zhiyuan and others},
  journal={arXiv preprint arXiv:2410.10594},
  year={2024}
}

@article{wang2023docllm,
  title={DocLLM: A layout-aware generative language model for multimodal document understanding},
  author={Wang, Dongsheng and Raman, Natraj and Sibue, Mathieu and Ma, Zhiqiang and Babkin, Petr and Kaur, Simerjot and Pei, Yulong and Nourbakhsh, Armineh and Liu, Xiaomo},
  journal={arXiv preprint arXiv:2401.00908},
  year={2023}
}

@article{dao2023flashattention,
  title={Flashattention-2: Faster attention with better parallelism and work partitioning},
  author={Dao, Tri},
  journal={arXiv preprint arXiv:2307.08691},
  year={2023}
}

@article{xiao2023efficient,
  title={Efficient streaming language models with attention sinks},
  author={Xiao, Guangxuan and Tian, Yuandong and Chen, Beidi and Han, Song and Lewis, Mike},
  journal={arXiv preprint arXiv:2309.17453},
  year={2023}
}

@article{chen2024far,
  title={How far are we to gpt-4v? closing the gap to commercial multimodal models with open-source suites},
  author={Chen, Zhe and Wang, Weiyun and Tian, Hao and Ye, Shenglong and Gao, Zhangwei and Cui, Erfei and Tong, Wenwen and Hu, Kongzhi and Luo, Jiapeng and Ma, Zheng and others},
  journal={arXiv preprint arXiv:2404.16821},
  year={2024}
}

@article{dai2024ligerkernel,
  title={Liger Kernel: Efficient Triton Kernels for LLM Training},
  author={Dai, Yun and Kothapalli, Vignesh and Song, Qingquan and Tang, Shao and Zhu, Siyu and Shimizu, Steven and Sahni, Shivam and Ning, Haowen and Chen, Yanning and others},
  journal={arXiv preprint arXiv:2410.10989},
  year={2024}
}

@misc{InternVL2,
  author = {OpenGVLab Team},
  title = {InternVL2: Better than the Best—Expanding Performance Boundaries of Open-Source Multimodal Models with the Progressive Scaling Strategy},
  year = {2024},
  url = {https://internvl.github.io/blog/2024-07-02-InternVL-2.0/}
}

@article{hu2024mplugdocowl2,
  title={mplug-docowl2: High-resolution compressing for ocr-free multi-page document understanding},
  author={Hu, Anwen and Xu, Haiyang and Zhang, Liang and Ye, Jiabo and Yan, Ming and Zhang, Ji and Jin, Qin and Huang, Fei and Zhou, Jingren},
  journal={arXiv preprint arXiv:2409.03420},
  year={2024}
}

@article{huang2024minimonkey,
  title={Mini-monkey: Alleviate the sawtooth effect by multi-scale adaptive cropping},
  author={Huang, Mingxin and Liu, Yuliang and Liang, Dingkang and Jin, Lianwen and Bai, Xiang},
  journal={arXiv preprint arXiv:2408.02034},
  year={2024}
}

@article{alayrac2022flamingo,
  title={Flamingo: a visual language model for few-shot learning},
  author={Alayrac, Jean-Baptiste and Donahue, Jeff and Luc, Pauline and Miech, Antoine and Barr, Iain and Hasson, Yana and Lenc, Karel and Mensch, Arthur and Millican, Katherine and Reynolds, Malcolm and others},
  journal={NeurIPS},
  volume={35},
  pages={23716--23736},
  year={2022}
}

@misc{liu2024llavanext,
    title={LLaVA-NeXT: Improved reasoning, OCR, and world knowledge},
    url={https://llava-vl.github.io/blog/2024-01-30-llava-next/},
    author={Liu, Haotian and Li, Chunyuan and Li, Yuheng and Li, Bo and Zhang, Yuanhan and Shen, Sheng and Lee, Yong Jae},
    month={January},
    year={2024}
}

@article{dao2022flashattention,
  title={Flashattention: Fast and memory-efficient exact attention with io-awareness},
  author={Dao, Tri and Fu, Dan and Ermon, Stefano and Rudra, Atri and R{\'e}, Christopher},
  journal={NeurIPS},
  volume={35},
  pages={16344--16359},
  year={2022}
}

@article{li2022paddleocr,
  title={PP-OCRv3: More attempts for the improvement of ultra lightweight OCR system},
  author={Li, Chenxia and Liu, Weiwei and Guo, Ruoyu and Yin, Xiaoting and Jiang, Kaitao and Du, Yongkun and Du, Yuning and Zhu, Lingfeng and Lai, Baohua and Hu, Xiaoguang and others},
  journal={arXiv preprint arXiv:2206.03001},
  year={2022}
}

@article{ye2023mplugdocowl,
  title={mplug-docowl: Modularized multimodal large language model for document understanding},
  author={Ye, Jiabo and Hu, Anwen and Xu, Haiyang and Ye, Qinghao and Yan, Ming and Dan, Yuhao and Zhao, Chenlin and Xu, Guohai and Li, Chenliang and Tian, Junfeng and others},
  journal={arXiv preprint arXiv:2307.02499},
  year={2023}
}

@article{hu2024mplug_docowl_1_5,
  title={mPLUG-DocOwl 1.5: Unified Structure Learning for OCR-free Document Understanding},
  author={Hu, Anwen and Xu, Haiyang and Ye, Jiabo and Yan, Ming and Zhang, Liang and Zhang, Bo and Li, Chen and Zhang, Ji and Jin, Qin and Huang, Fei and others},
  journal={arXiv preprint arXiv:2403.12895},
  year={2024}
}

@misc{ma2024mmlong,
      title={MMLongBench-Doc: Benchmarking Long-context Document Understanding with Visualizations}, 
      author={Yubo Ma and Yuhang Zang and Liangyu Chen and Meiqi Chen and Yizhu Jiao and Xinze Li and Xinyuan Lu and Ziyu Liu and Yan Ma and Xiaoyi Dong and Pan Zhang and Liangming Pan and Yu-Gang Jiang and Jiaqi Wang and Yixin Cao and Aixin Sun},
      year={2024},
      eprint={2407.01523},
      archivePrefix={arXiv},
      primaryClass={cs.CV},
      url={https://arxiv.org/abs/2407.01523}, 
}

@article{wang2024needle,
  title={Needle In A Multimodal Haystack}, 
  author={Wang, Weiyun and Zhang, Shuibo and Ren, Yiming and Duan, Yuchen and Li, Tiantong and Liu, Shuo and Hu, Mengkang and Chen, Zhe and Zhang, Kaipeng and Lu, Lewei and Zhu, Xizhou and Luo, Ping and Qiao, Yu and Dai, Jifeng and Shao, Wenqi and Wang, Wenhai},
  journal={arXiv preprint arXiv:2406.07230},
  year={2024}
}

@article{cho2024m3docrag,
  title={M3DocRAG: Multi-modal Retrieval is What You Need for Multi-page Multi-document Understanding},
  author={Cho, Jaemin and Mahata, Debanjan and Irsoy, Ozan and He, Yujie and Bansal, Mohit},
  journal={arXiv preprint arXiv:2411.04952},
  year={2024}
}

@inproceedings{duan2025docopilot,
  title={Docopilot: Improving Multimodal Models for Document-Level Understanding},
  author={Duan, Yuchen and Chen, Zhe and Hu, Yusong and Wang, Weiyun and Ye, Shenglong and Shi, Botian and Lu, Lewei and Hou, Qibin and Lu, Tong and Li, Hongsheng and others},
  booktitle={Proceedings of the Computer Vision and Pattern Recognition Conference},
  pages={4026--4037},
  year={2025}
}

@article{idefics,
  author       = {Hugo Lauren{\c{c}}on and
                  L{\'{e}}o Tronchon and
                  Matthieu Cord and
                  Victor Sanh},
  title        = {What matters when building vision-language models?},
  journal      = {CoRR},
  volume       = {abs/2405.02246},
  year         = {2024}
}

@article{longva,
  author       = {Peiyuan Zhang and
                  Kaichen Zhang and
                  Bo Li and
                  Guangtao Zeng and
                  Jingkang Yang and
                  Yuanhan Zhang and
                  Ziyue Wang and
                  Haoran Tan and
                  Chunyuan Li and
                  Ziwei Liu},
  title        = {Long Context Transfer from Language to Vision},
  journal      = {CoRR},
  volume       = {abs/2406.16852},
  year         = {2024}
}

@article{llava-next-interleave,
  author       = {Feng Li and
                  Renrui Zhang and
                  Hao Zhang and
                  Yuanhan Zhang and
                  Bo Li and
                  Wei Li and
                  Zejun Ma and
                  Chunyuan Li},
  title        = {LLaVA-NeXT-Interleave: Tackling Multi-image, Video, and 3D in Large
                  Multimodal Models},
  journal      = {CoRR},
  volume       = {abs/2407.07895},
  year         = {2024}
}

@inproceedings{feng2021optimal,
  title={Optimal gradient checkpoint search for arbitrary computation graphs},
  author={Feng, Jianwei and Huang, Dong},
  booktitle={Proceedings of the IEEE/CVF Conference on Computer Vision and Pattern Recognition},
  pages={11433--11442},
  year={2021}
}

@inproceedings{tang2025scaling,
  title={Scaling On-Device GPU Inference for Large Generative Models},
  author={Tang, Jiuqiang and Sorokin, Raman and Ignasheva, Ekaterina and Jensen, Grant and Chen, Lin and Lee, Juhyun and Kulik, Andrei and Grundman, Matthias},
  booktitle={Proceedings of the Computer Vision and Pattern Recognition Conference},
  pages={6355--6364},
  year={2025}
}

@article{cui2025paddleocr,
  title={Paddleocr 3.0 technical report},
  author={Cui, Cheng and Sun, Ting and Lin, Manhui and Gao, Tingquan and Zhang, Yubo and Liu, Jiaxuan and Wang, Xueqing and Zhang, Zelun and Zhou, Changda and Liu, Hongen and others},
  journal={arXiv preprint arXiv:2507.05595},
  year={2025}
}

@misc{loshchilov2019decoupledweightdecayregularization,
      title={Decoupled Weight Decay Regularization}, 
      author={Ilya Loshchilov and Frank Hutter},
      year={2019},
      eprint={1711.05101},
      archivePrefix={arXiv},
      primaryClass={cs.LG},
      url={https://arxiv.org/abs/1711.05101}, 
}

@inproceedings{newsvideoqa,
  author       = {Soumya Jahagirdar and
                  Minesh Mathew and
                  Dimosthenis Karatzas and
                  C. V. Jawahar},
  title        = {Watching the News: Towards VideoQA Models that can Read},
  booktitle    = {{WACV}},
  pages        = {4430--4439},
  publisher    = {{IEEE}},
  year         = {2023}
}

@article{mpdocvqa,
  author       = {Rub{\`{e}}n Tito and
                  Dimosthenis Karatzas and
                  Ernest Valveny},
  title        = {Hierarchical multimodal transformers for Multi-Page DocVQA},
  journal      = {CoRR},
  volume       = {abs/2212.05935},
  year         = {2022}
}

@inproceedings{dude,
  author       = {Jordy Van Landeghem and
                  Rafal Powalski and
                  Rub{\`{e}}n Tito and
                  Dawid Jurkiewicz and
                  Matthew B. Blaschko and
                  Lukasz Borchmann and
                  Micka{\"{e}}l Coustaty and
                  Sien Moens and
                  Michal Pietruszka and
                  Bertrand Anckaert and
                  Tomasz Stanislawek and
                  Pawel J{\'{o}}ziak and
                  Ernest Valveny},
  title        = {Document Understanding Dataset and Evaluation {(DUDE)}},
  booktitle    = {{ICCV}},
  pages        = {19471--19483},
  publisher    = {{IEEE}},
  year         = {2023}
}

@article{ma2024mmlongbench,
  title={Mmlongbench-doc: Benchmarking long-context document understanding with visualizations},
  author={Ma, Yubo and Zang, Yuhang and Chen, Liangyu and Chen, Meiqi and Jiao, Yizhu and Li, Xinze and Lu, Xinyuan and Liu, Ziyu and Ma, Yan and Dong, Xiaoyi and others},
  journal={Advances in Neural Information Processing Systems},
  volume={37},
  pages={95963--96010},
  year={2024}
}

@inproceedings{biten2019scene,
  title={Scene text visual question answering},
  author={Biten, Ali Furkan and Tito, Rub{\`e}n and Mafla, Andres and Gomez, Lluis and Karatzas, Dimosthenis},
  booktitle={ICDAR},
  year={2019}
}

@inproceedings{blau2024gram,
  title={GRAM: Global reasoning for multi-page VQA},
  author={Blau, Tsachi and Fogel, Sharon and Ronen, Roi and Golts, Alona and Ganz, Roy and Ben Avraham, Elad and Aberdam, Aviad and Tsiper, Shahar and Litman, Ron},
  booktitle={Proceedings of the IEEE/CVF Conference on Computer Vision and Pattern Recognition},
  pages={15598--15607},
  year={2024}
}

@inproceedings{li2023blip,
  title={Blip-2: Bootstrapping language-image pre-training with frozen image encoders and large language models},
  author={Li, Junnan and Li, Dongxu and Savarese, Silvio and Hoi, Steven},
  booktitle={International conference on machine learning},
  pages={19730--19742},
  year={2023},
  organization={PMLR}
}

@article{li2024tokenpacker,
  title={Tokenpacker: Efficient visual projector for multimodal llm},
  author={Li, Wentong and Yuan, Yuqian and Liu, Jian and Tang, Dongqi and Wang, Song and Zhu, Jianke and Zhang, Lei},
  journal={arXiv preprint arXiv:2407.02392},
  year={2024}
}

@article{hu2024mplug,
  title={mplug-docowl2: High-resolution compressing for ocr-free multi-page document understanding},
  author={Hu, Anwen and Xu, Haiyang and Zhang, Liang and Ye, Jiabo and Yan, Ming and Zhang, Ji and Jin, Qin and Huang, Fei and Zhou, Jingren},
  journal={arXiv preprint arXiv:2409.03420},
  year={2024}
}

@inproceedings{biten2022latr,
  title={Latr: Layout-aware transformer for scene-text vqa},
  author={Biten, Ali Furkan and Litman, Ron and Xie, Yusheng and Appalaraju, Srikar and Manmatha, R},
  booktitle={Proceedings of the IEEE/CVF conference on computer vision and pattern recognition},
  pages={16548--16558},
  year={2022}
}

@inproceedings{ganz2023towards,
  title={Towards models that can see and read},
  author={Ganz, Roy and Nuriel, Oren and Aberdam, Aviad and Kittenplon, Yair and Mazor, Shai and Litman, Ron},
  booktitle={Proceedings of the IEEE/CVF international conference on computer vision},
  pages={21718--21728},
  year={2023}
}

@inproceedings{ye2023deepsolo,
  title={DeepSolo: Let Transformer Decoder with Explicit Points Solo for Text Spotting},
  author={Ye, Maoyuan and Zhang, Jing and Zhao, Shanshan and Liu, Juhua and Liu, Tongliang and Du, Bo and Tao, Dacheng},
  booktitle={Proceedings of the IEEE/CVF Conference on Computer Vision and Pattern Recognition},
  pages={19348--19357},
  year={2023}
}

@article{di2025streaming,
  title={Streaming video question-answering with in-context video kv-cache retrieval},
  author={Di, Shangzhe and Yu, Zhelun and Zhang, Guanghao and Li, Haoyuan and Zhong, Tao and Cheng, Hao and Li, Bolin and He, Wanggui and Shu, Fangxun and Jiang, Hao},
  journal={arXiv preprint arXiv:2503.00540},
  year={2025}
}

@misc{Executio91:online,
author = {},
  title = {Execution Providers | onnxruntime},
  url = "https://onnxruntime.ai/docs/execution-providers/",
month = {},
year = {},
  note = "[Online; accessed 2025-11-20]"
}

@misc{Qualcomm95:online,
author = {},
  title = {Qualcomm - QNN | onnxruntime},
  url = "https://onnxruntime.ai/docs/execution-providers/QNN-ExecutionProvider.html",
month = {},
year = {},
  note = "[Online; accessed 2025-11-20]"
}

@misc{Buy138in63:online,
author = {},
  title = {Buy 13.8-inch Surface Laptop, Copilot+ PC with Windows - Microsoft Store},
  url = "https://www.microsoft.com/en-gb/store/configure/surface-laptop-13-8-inch/8mzbmmcjzqjk",
month = {},
year = {},
  note = "[Online; accessed 2025-11-21]"
}

@inproceedings{dettmers2023qlora,
  title={QLoRA: Efficient Finetuning of Quantized Large Language Models},
  author={Dettmers, Tim and others},
  booktitle={NeurIPS},
  year={2023}
}
}

\end{document}